\documentclass[sigconf]{acmart}

\AtBeginDocument{%
  \providecommand\BibTeX{{%
    \normalfont B\kern-0.5em{\scshape i\kern-0.25em b}\kern-0.8em\TeX}}}

\copyrightyear{2024}
\acmYear{2024}
\setcopyright{acmlicensed}\acmConference[CHI '24]{Proceedings of the CHI Conference on Human Factors in Computing Systems}{May 11--16, 2024}{Honolulu, HI, USA}
\acmBooktitle{Proceedings of the CHI Conference on Human Factors in Computing Systems (CHI '24), May 11--16, 2024, Honolulu, HI, USA}
\acmDOI{10.1145/3613904.3642790}
\acmISBN{979-8-4007-0330-0/24/05}

\usepackage{multirow}
\usepackage{diagbox}
\usepackage[utf8]{inputenc} 
\usepackage{graphicx}
\usepackage{stfloats}

\usepackage{balance}       
\usepackage{graphics}      
\usepackage{color}
\usepackage{booktabs}
\usepackage{textcomp}
\usepackage{subcaption}
\usepackage{enumerate}
\usepackage{makecell}
\usepackage{multicol}
\usepackage{multirow}
\usepackage{array}
\usepackage{fdsymbol}
\usepackage{enumitem}
\usepackage{amsmath}
\usepackage{arydshln}
\usepackage{stfloats}
\usepackage{graphicx}
\usepackage{amsthm}
\usepackage{listings}
\usepackage{caption} 
\usepackage{xspace}
\usepackage[linesnumbered]{algorithm2e}
\usepackage{csquotes}
\usepackage[bottom]{footmisc}
\usepackage{textcomp}

\newcommand*{\eg}{\textit{e.g.},\xspace}
\newcommand*{\ie}{\textit{i.e.},\xspace}

\newenvironment{s_itemize}{
\begin{itemize}[leftmargin=*]
  \setlength{\itemsep}{3pt}
  \setlength{\parskip}{0pt}
  \setlength{\parsep}{0pt}
}{\end{itemize}}

\definecolor{DarkGreen}{HTML}{5DAC81}
\definecolor{myred}{RGB}{255,0,0}

\newcommand\review[1]{\textcolor{black}{#1}}
\newcommand\minorreview[1]{\textcolor{black}{#1}}

\setcitestyle{nocompress}

\begin{document}

\title{MindShift: Leveraging Large Language Models for Mental-States-Based Problematic Smartphone Use Intervention}

\author{Ruolan Wu}
\affiliation{%
  \institution{Tsinghua University}
  \country{Beijing, Beijing, China}
}
\email{wurl21@mails.tsinghua.edu.cn}

\author{Chun Yu}
\affiliation{%
  \institution{Tsinghua University}
  \country{Beijing, Beijing, China}
  }
\email{chunyu@tsinghua.edu.cn}
\authornote{Corresponding author.}

\author{Xiaole Pan}
\affiliation{%
  \institution{Tsinghua University}
  \country{Beijing, Beijing, China}
}
\email{pxl22@mails.tsinghua.edu.cn}

\author{Yujia Liu}
\affiliation{%
  \institution{Tsinghua University}
  \country{Beijing, Beijing, China}
}
\email{l-yj22@mails.tsinghua.edu.cn}

\author{Ningning Zhang}
\affiliation{%
  \institution{Tsinghua University}
  \country{Beijing, Beijing, China}
}
\email{znn18@tsinghua.org.cn}

\author{Yue Fu}
\affiliation{%
  \institution{University of Washington}
  \country{Seattle, Washington, USA}
}
\email{chrisfu@uw.edu}

\author{Yuhan Wang}
\affiliation{%
  \institution{Beijing University of Posts and Telecommunications}
  \country{Beijing, Beijing, China}
}
\email{2020211730@bupt.cn}

\author{Zhi Zheng}
\affiliation{%
  \institution{Tsinghua University}
  \country{Beijing, Beijing, China}
}
\email{georgezhengzhi@gmail.com}

\author{Li Chen}
\affiliation{%
  \institution{Tsinghua University}
  \country{Beijing, Beijing, China}
}
\email{chenli19@mails.tsinghua.edu.cn}

\author{Qiaolei Jiang}
\affiliation{%
  \institution{Tsinghua University}
  \country{Beijing, Beijing, China}
}
\email{qiaoleijiang@tsinghua.edu.cn}

\author{Xuhai Xu}
\affiliation{%
  \institution{Massachusetts Institute of Technology}
    \country{Cambridge, MA, USA}
}
\email{xoxu@mit.edu}

\author{Yuanchun Shi}
\affiliation{%
  \institution{Tsinghua University}
  \country{Beijing, Beijing, China}
}
\email{shiyc@tsinghua.edu.cn}

\renewcommand{\shortauthors}{Wu et al.}
\renewcommand{\shorttitle}{MindShift}


\begin{abstract}

Problematic smartphone use negatively affects physical and mental health.
Despite the wide range of prior research, existing persuasive techniques are not flexible enough to provide dynamic persuasion content based on users' physical contexts and mental states.
We first conducted a Wizard-of-Oz study (N=12) and an interview study (N=10) to summarize the mental states behind problematic smartphone use: boredom, stress, and inertia.
This informs our design of four persuasion strategies: understanding, comforting, evoking, and scaffolding habits.
We leveraged large language models (LLMs) to enable the automatic and dynamic generation of effective persuasion content.
We developed MindShift, a novel LLM-powered problematic smartphone use intervention technique.
MindShift takes users' in-the-moment app usage behaviors, physical contexts, mental states, goals \& habits as input, and generates personalized and dynamic persuasive content with appropriate persuasion strategies.
We conducted a 5-week field experiment (N=25) to compare MindShift \review{with its simplified version (remove mental states) and baseline techniques (fixed reminder)}.
The results show that MindShift \review{improves intervention acceptance rates by 4.7-22.5\% and reduces smartphone usage duration by 7.4-9.8\%.}
Moreover, users have a significant drop in smartphone addiction scale scores and a rise in self-efficacy scale scores.
Our study sheds light on the potential of leveraging LLMs for context-aware persuasion in other behavior change domains.

\end{abstract}



\maketitle

\section{Introduction}

\begin{figure*}
    \centering
    \includegraphics[width=0.9\linewidth]{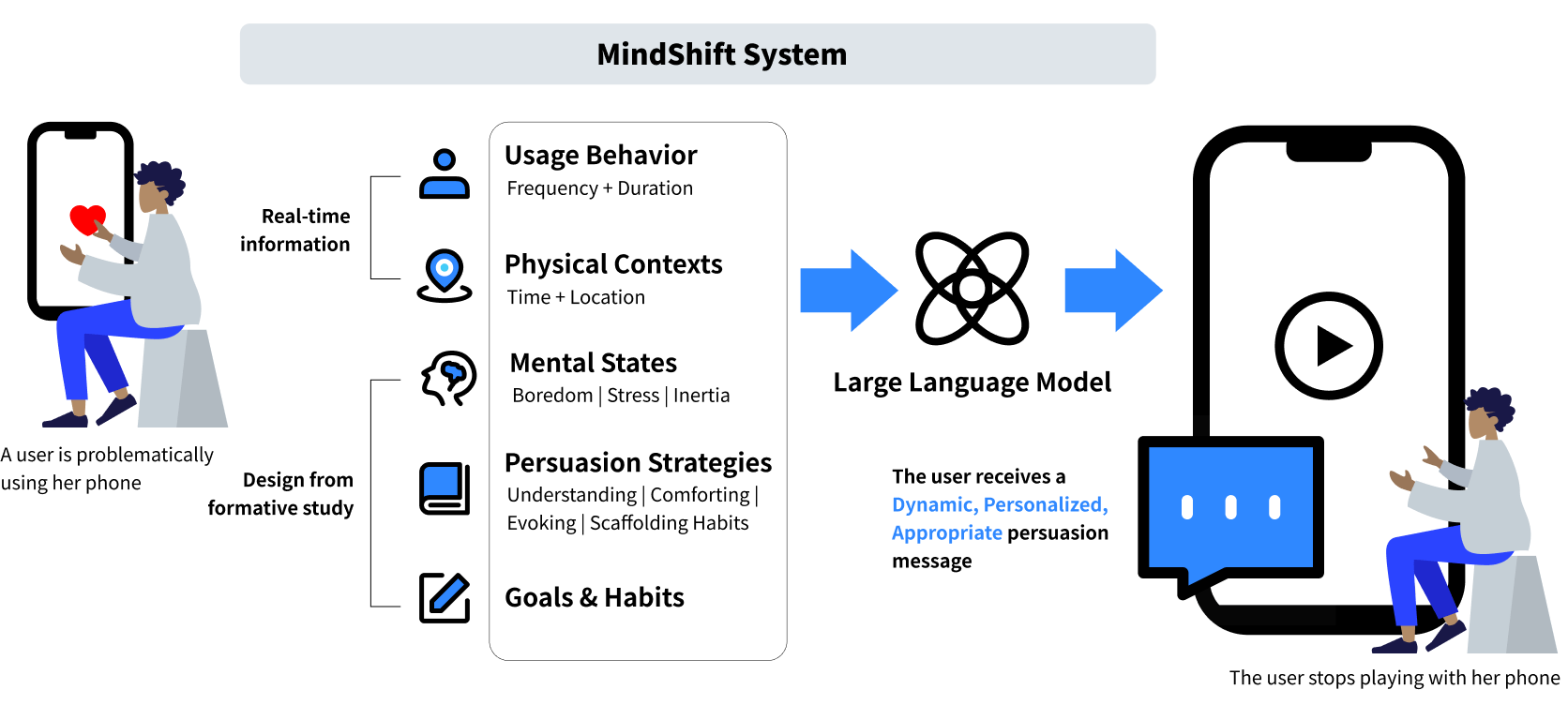}
    \caption{Overview of MindShift. When users exhibit problematic phone usage, MindShift actively collects data on phone usage behavior, physical contexts, and mental states, and uses the customized persuasion strategies we designed along with users' goals and habits, to generate prompts. Then, the large language model will generate a persuasion message. Finally, the persuasion will show up in users' phones to encourage more mindful usage.}
    \Description{(Left) A picture illustrating that a user is problematically using her phone. (Medium) A flowchart illustrating the MindShift system. The left side of the flowchart is the system's inputs, organized vertically into two main sections. The top section, labeled 'Real-time information,' details two factors: (1) 'Usage Behavior' (containing frequency and duration of phone usage), and (2)'Physical Context' (describing the time and location of the user). Below this, the second section is titled 'Design from Formative Study.' It includes three variables: (1)'Mental State' (options are Stress, Boredom, or Inertia), (2)'Persuasion Strategy' (options are Understanding, Comforting, Evoking, or Scaffolding Habits), and (3)'Goals and habits.' An arrow links these inputs to an icon representing a 'Large Language Model'. On the right side of the flowchart, a depiction shows a user receiving a personalized message from the system. (Right) A picture illustrating that the user stops playing with her phone.}
    \label{fig:overview}
\end{figure*}

In recent years, the ubiquitous presence of smartphones has increased people's reliance on digital devices, resulting in problematic smartphone usage behaviors, \review{\ie excessive or mindless usage with negative consequences~\cite{lukoff2018makes,hwang2012smartphone}},
especially among adolescents and young adults~\cite{lee2014hooked}.
Prior studies suggest that problematic smartphone usage can detrimentally affect people in various areas such as efficiency (leading to diminished academic or work performance~\cite{arefin2018impact,duke2017smartphone}), physical well-being (resulting in decreased sleep and activity levels~\cite{clayton2015extended,lapointe2013smartphone, xu2019leveraging, xu2021leveraging}), and mental health (manifesting as anxiety and depression~\cite{hartanto2016smartphone,harwood2014constantly, xu_globem_2022}).
Many individuals have recognized their problematic smartphone usage and sought to reduce their over-reliance on smartphones~\cite{hiniker2016mytime,ko2015nugu}.

A plethora of academic research and commercial products provide just-in-time (JIT) smartphone use interventions, intervening precisely when problematic use occurs~\cite{nahum2018just}.
These interventions fall into four categories based on their enforcement levels:
(1) 
\review{Self-monitoring:}
offering insights on phone usage patterns \review{via notification or visualization}, enhancing user awareness about their smartphone habits~\cite{whittaker2016don,kim2016timeaware,collins2014barriers,andone2016menthal};
(2) Reminders: countering immediate phone indulgence and promoting self-reflection through pop-up notifications~\cite{hiniker2016mytime,park2015initiating};
(3) Interaction friction: raising the effort needed to use the phone, thereby reducing its allure by introducing tasks like typing~\cite{park2018interaction,okeke2018good,xu2022typeout};
(4) Lockout: disabling the user's phone access for a specified duration~\cite{kim2019goalkeeper,ko2015nugu,lochtefeld2013appdetox,kim2019lockntype}.
However, there are a few gaps among existing intervention techniques. 

\review{
First, existing methods are limited to strike a balance between effective intervention engagement and good usability~\cite{monge2019race}. The first two categories rely on the user's self-control and are easily ignored, leading to low engagement and limited effectiveness~\cite{whittaker2016don,kim2016timeaware,hiniker2016mytime}. The other two types 
are more restrictive, often causing user frustration due to reduced usability across different contexts~\cite{kim2019goalkeeper}. Our approach utilizes reminder-based interventions with persuasive content to encourage reduced smartphone usage. Persuasion, typically through natural language to influence people's thoughts and behavior~\cite{simons1976persuasion,rapp2002aristotle,cialdini2007influence}, is more effective with diverse and context-specific content~\cite{kaptein2015personalizing, kaptein2012adaptive}, as supported by recent studies highlighting the success of personalized~\cite{kovacs2018rotating, choe2017semi,harrison2015activity, kaptein2012adaptive}, context-aware~\cite{kim2017let,ko2015nugu,thomas2021systematic,purohit2019functional} interventions. However, most current reminders use repetitive, template-based content, reducing efficacy~\cite{apple,Tiktok,xu2022typeout}. To overcome this, we employ Large Language Models (LLMs)~\cite{OpenAI,chowdhery2022palm} to generate varied persuasive content. LLMs' reasoning ability provides a promising solution to infer users' current activities based on contextual information collected from smartphone sensors, such as time and location~\cite{chen2023gap,kim2024health,englhardt2023classification}, enabling the creation of more relevant and effective intervention language.
}

\review{Second, we identified the opportunity to leverage mental states associated with problematic smartphone use, an essential aspect of user contexts. While some current interventions use context-based strategies, like triggering interventions at specific times and locations~\cite{kim2017let,ko2015nugu}, they tend to focus mainly on external physical contexts, neglecting internal mental factors. Mental factors like stress and negative emotions are increasingly recognized as one key factor leading to problematic smartphone use ~\cite{lyngs2019self,duvenage2020technology,wang2014understanding,wang2020we}. However, existing studies primarily address prolonged mental states rather than momentary contexts.
Our study aims to bridge this gap by integrating an understanding of in-the-moment mental states into the intervention framework. We believe that a more holistic approach, considering both the physical and mental contexts, could enhance intervention effectiveness.}

To address these gaps, we first conducted a Wizard-of-Oz study (N=12), followed by an interview study (N=10) to better understand users' mental states during problematic phone usage.
Focusing on habitual usage (\ie ritualistic behavior, without a clear goal, such as passive social media content consumption)~\cite{rubin2009uses,hiniker2016would}, we summarized three major mental states to address: \textit{boredom}, \textit{stress}, and \textit{inertia}. Building on the Dual Systems Theory~\cite{hofmann2009impulse} and the ERG (Existence, Relatedness, and Growth) Theory~\cite{caulton2012development}, we proposed four persuasion strategies: 1) understanding, 2) comforting, 3) evoking, and 4) scaffolding habits.

Integrating our persuasion strategies with LLMs, we designed and implemented MindShift (Figure \ref{fig:overview}), a new JIT intervention technique that can provide dynamic, personalized persuasion content based on user contexts.
MindShift leverages LLMs' strong capability in commonsense comprehension and natural language generation~\cite{chowdhery2022palm,min2021recent} to generate proper and effective persuasive content based on real-time information (phone usage behavior, physical contexts, and mental states) and long-term user states (user goals and habits), guided by persuasion strategies we designed.

To evaluate MindShift's effectiveness, we conducted a 5-week field experiment deploying our intervention to 25 participants.
We compared MindShift against the baseline, a basic notification-based intervention that asked users to reflect on and report the purpose of their smartphone usage.
Moreover, to assess the effect of the mental states factor, we compared MindShift against a simplified version, MindShift-Simple, that excludes the mental states factor from the LLM-based content generation.

\review{Our study results indicate that MindShift and MindShift-Simple outperformed the baseline method on the intervention acceptance rate by 22.5\% and 17.8\%, respectively, with statistical significance. They also significantly reduce overall app opening frequency by 12.1\% and 14.4\% and app usage duration by 9.8\% and 2.4\%.
Comparing MindShift and MindShift-Simple, including the mental state factor enhances the persuasion acceptance rate by 8.1\% with statistical significance}.
Moreover, the subjective report data shows that participants using MindShift and MindShift-Simple experience a significant reduction in smartphone addiction scale (SAS) score (34.7\% and 25.8\% respectively) and an increase in the self-efficacy scale score (10.7\% and 10.4\% respectively).

Our paper makes the following contributions:
\begin{enumerate}
\item We conducted a Wizard-of-Oz study and an interview study, uncovering three major mental states (\textit{boredom}, \textit{stress}, and \textit{inertia)} during habitual smartphone use, which led us to design four persuasion strategies grounded in the Dual Systems Theory and the ERG Theory: \textit{Understanding, Comforting, Evoking, }and\textit{ Scaffolding Habits.}
\item We created MindShift, a novel persuasive intervention technique leveraging LLMs to generate dynamic and personalized persuasion content based on users' phone usage behavior, physical contexts, mental states, goals and habits, and appropriate persuasion strategies.
\item We conducted a field experiment by deploying MindShift, demonstrating significant improvements in intervention acceptance rates and reduced smartphone use by Mindshift. Users' subjective feedback also corroborated these observations, validating the effectiveness of MindShift.
\end{enumerate}

\section{Related work}
In this section, we define problematic smartphone use and habitual smartphone use (Sec. \ref{sub:related-definition}), \review{explore the reasons behind engagement in problematic smartphone use} (Sec. \ref{sub:related-tehory}). We then briefly overview existing intervention techniques (Sec. \ref{sub:related-intervention}). Finally, close to our work, we introduce behavior change persuasion techniques and their relationship with the emergence of LLMs (Sec. \ref{sub:related-persuasion}). 

\subsection{Problematic Smartphone Use}
\label{sub:related-definition}
Many studies have explored the definition of problematic smartphone use, which can be broadly classified into two categories.
The first category defines whether users exhibit addictive behaviors toward their phones. Some studies assess addictive behavior by measuring the level of user dependence on smartphones through questionnaires, such as the Smartphone Addiction Scale (SAS) and the Smartphone Addiction Inventory (SPAI)~\cite{lin2014development,kwon2013development}.
The second category defines whether a specific instance of phone use is problematic. Growing research suggests that problematic smartphone use is determined not only by excessive use but also by the purpose and content of use in specific situations~\cite{harwood2014constantly,prasad2021addressing,purohit2020designing,lukoff2023switchTube,lukoff2021design,lukoff2018makes}.
Studies have indicated that phone use purposes can be categorized into (1) habitual use that is performed unconsciously and ritualistically usually without a specific goal~\cite{pinder2018digital}, and (2) instrumental use with a specific task or goal in mind~\cite{lukoff2018makes}.
Existing research suggests that habitual use should be the primary target for intervention~\cite{morris2018towards,lukoff2018makes,pinder2018digital}.
In this paper, we use SAS to measure users' level of addictive smartphone usage.
We also distinguish users' phone use purpose and focus on intervening habitual use.

\subsection{\review{Understanding Problematic Smartphone Use through A Dual Systems Perspective}}
\label{sub:related-tehory}
Understanding what leads to problematic smartphone use is essential for the design of effective persuasion strategies.
The Dual Systems Theory~\cite{hofmann2009impulse,kahneman2011thinking} has been used to explain the phone usage patterns~\cite{pinder2018digital}. This theory divides human cognitive activities into two types: System 1 (fast, intuitive, unconscious) and System 2 (slow, analytical, conscious). Problematic smartphone use is typically driven by System 1~\cite{hagger2016non}, as it mainly involves unconscious, rapid responses and is easy to be guided by instant gratification.
Research suggests that two key factors contribute to the failure of users to act on their goals: (1) limited ability of System 2 control; and (2) fluctuations of System 2 caused by emotional states and fatigue~\cite{lyngs2019self}. \review{For the first factor, a growing amount of research suggests that the limited ability of control is attributed more to apps' deliberate design than to users themselves \cite{purohit2023starving,baughan2022don,lukoff2021design, monge2023defining,gray2018dark}. Recent work identifies types of attention-capture deceptive designs in digital interfaces, such as neverending autoplay and infinite scroll \cite{monge2023defining}. For the second factor,} some findings suggest that mental states play a significant role in habitual smartphone use \cite{pinder2018digital,wang2014understanding,wang2020we} and previous research has identified external contexts, \review{such as social
awkwardness~\cite{tran2019modeling,purohit2020designing}, that may trigger the habitual use. However, there is limited research exploring what specific kinds of users' in-the-moment mental states behind habitual use}.
In our work, we pinpoint the major mental states linked to habitual use and propose corresponding persuasion strategies.

\subsection{Problematic Smartphone Use Intervention}
\label{sub:related-intervention}
\review{Existing problematic smartphone use intervention techniques fall into two groups: external interventions that monitor and limit use, and internal interventions that change the interface itself.}

\review{External intervention can be} roughly divided into four categories based on enforcement level:
The first category provides users with information about their behavior such as visualization of usage~\cite{whittaker2016don,kim2016timeaware,hiniker2016mytime,collins2014barriers,apple, lyngs2022goldilocks,nwagu2023design,orzikulova2024time2stop,lu2024interactout}, requiring users to view it themselves to increase awareness of phone usage. The second actively sends reminders to users to provoke their reflection such as reminding users of their daily goals~\cite{hiniker2016mytime,lyngs2020just}, informing their usage time~\cite{apple} or the number of opens~\cite{purohit2023designing}. This category presents text to users and, therefore, serves as a persuasion. The third involves increasing the difficulty of using the phone and intentionally slowing down user interactions to suppress the desire to use it, such as requiring users to enter random numbers or type self-reflective text~\cite{park2018interaction,xu2022typeout} and keeping the phone vibrating continuously~\cite{okeke2018good}.
The fourth is particularly forceful by directly locking the users' apps or phones for a specific duration~\cite{forest,kim2019goalkeeper,ko2015nugu}.
There are concerns about these methods' ability to strike an optimal balance between usability and effectiveness~\cite{monge2019race}. 

\review{Internal intervention involves redesigning app interfaces to counteract attention-capturing deceptive designs~\cite{monge2023defining}. 
For example, increasing user awareness of time spent through reading history labels ~\cite{baughan2022don} and specific color change ~\cite{monge2023nudging}, eliminating the addictive design of infinite scroll through removing ~\cite{lyngs2020just} and adjusting the newsfeed ~\cite{kim2016timeaware,purohit2023starving,zhang2022monitoring}, decreasing the guilty pleasure recommendations through using adaptable commitment interface ~\cite{lukoff2023switchTube} and redesigning search interface ~\cite{monge2023nudging}.} 
\review{Compared to external intervention, internal intervention can better balance effectiveness and long-term experience~\cite{zhang2022monitoring,purohit2023starving}. However, these internal methods often require third-party development, as large companies rarely adopt such designs themselves due to financial interests~\cite{gray2018dark}. This necessitates additional development costs and the proposed design is typically tailored for a single app, making it hard to apply broadly.}

\review{Therefore, we hope to create a universal external intervention, using the form of reminders to ensure usability while boosting intervention effectiveness through \minorreview{personalized persuasion.}}

\subsection{Persuasion for User Behavior Change and Large Language Models}
\label{sub:related-persuasion}
Persuasion is a psychological approach designed to influence attitudes, beliefs, or behaviors \cite{cialdini2007influence}. Language is the most common means of persuasion~\cite{simons1976persuasion,rapp2002aristotle}, and leverages facts, emotional appeals, and so on to achieve its goal. Its effectiveness has been shown in multiple fields, such as advertising to encourage consumers to buy a product \cite{bernritter2017self}, supporting mental health such as coping with stress \cite{morris2018towards,orji2018persuasive}, and managing physical health such as reducing snacking behavior \cite{kaptein2012adaptive}.
For smartphone use intervention, persuasion usually appears as reminders, such as leveraging the user's usage time in a template format~\cite{apple,Tiktok}, or some thought-provoking statements~\cite{xu2022typeout}.
Prior work has suggested that personalizing content can enhance persuasion effectiveness such as reducing snacks ~\cite{kaptein2015personalizing, kaptein2012adaptive}.
Also, varying the timing and content of interventions, sometimes even randomly, can improve effectiveness. In contrast, static interventions tend to lose influence over time~\cite{kovacs2018rotating}.

The advent of Large Language Models (LLMs), like ChatGPT\cite{OpenAI} and PaLM\cite{chowdhery2022palm}, has made vast progress in personalized and diverse content generation.
Recent studies have explored various health applications supported by LLMs, such as health information seeking~\cite{yunxiang2023chatdoctor,mahmood2023llm,yang2023talk2care}, mental health support~\cite{lamichhane2023evaluation, kumar_2023_exploring,xu2023mentalllm}, personal health coaching~\cite{wei_leveraging_2023, montagna2023data}, health education~\cite{kung2023performance}, and public health interventions~\cite{jo2023understanding}. \review{These applications showcase LLMs' capabilities in knowledge delivery and emotional support. Compared to them, our study further explores LLMs for just-in-time behavior change and intervention, beyond information presentation.}
\section{Mental States of Habitual Smartphone Use and Persuasion Strategies}
\label{sec:formative_studies}

To comprehend the mental states of users' smartphone use and guide our intervention system design, we initiated a Wizard-of-Oz (WoZ) study, followed by a semi-structured interview study (Sec.~\ref{sub:formative_studies:studies}).
We summarized the main takeaways in Sec.~\ref{sub:formative_studies:mental_state}. \review{Based on theories and our findings, we devised four persuasion strategies and their implementation under different mental states}
(Sec.~\ref{sub:formative_studies:strategy}).

\subsection{Exploratory Wizard-of-Oz \& Semi-structured Interview Studies}
\label{sub:formative_studies:studies}

\review{To identify particular smartphone usage behaviors requiring targeted interventions, we first recruited 12 end-users (6 females and 6 males, aged 18-28) and conducted a 5-day WoZ study in the wild. The findings suggested ideas for persuasion content design. For deeper insights into participants' mental states and concrete intervention design materials, we recruited another group of 10 users (5 females and 5 males, aged 18-29) and conducted a semi-structured interview study\footnote{Since the two studies are close in time, we choose a separate set of participants for semi-structured interviews to avoid the impact of intervention in the WoZ study.}.
Both studies are approved by the institution's IRB.
Our studies focused on young adults who have been reported to have the most severe problematic smartphone use issues~\cite{mitchell2018predictors,lee2014hooked}. However, we do recognize that our sampling could limit the generalizability of our findings. We discuss this as a limitation in Sec.~\ref{sec:discussion}.
}

\subsubsection{Wizard-of-Oz Study}

We developed a chatbot system for smartphones that tracks user app activity. First, we asked participants to select apps for intervention, adding them to a blacklist. The chatbot then would send persuasive messages to the participants upon opening a blacklisted app. 

To reduce the observation effect, participants were told that it was an automatic chatbot instead of a human~\cite{kazdin1982observer}. In reality, when participants opened a blacklisted app, a human experimenter would receive an email notification. \review{Based on smartphone usage duration and frequency (see the detection method in Sec. \ref{sec:implementation}), the experimenter designed and delivered persuasive messages. Inspired by existing literature on persuasion design~\cite{ham2020learning,albarracin2018psychology,block1997effects}, our messages fell into 4 types (see examples in Table \ref{tab:woz-message} in Appendix): (1) usage notice, telling participants their usage data such as the accumulated usage time today and time since last use, (2) practical guidance, asking participants' goals today and suggesting tasks instead of smartphone use, (3) encouragement, praising and cheering participants to keep smartphones away, and (4) deterrent, alarming participants the consequences of using smartphones such as task delay and admonishing them to stop.}

\review{Every evening, researchers conducted a brief 15-minute online interview in person with each participant, structured around four questions: (1) What was your overall experience of using the chatbot? How did it change your smartphone usage? (2) Why were you using your phone at a particular time? (3) What were your reactions to the persuasive message, and why? (4) How did you like the persuasive message? How can it be improved?} 
At the study's conclusion, we informed participants that the chatbot was actually operated by a human experimenter at the back end.

\review{All persuasive messages and participants' responses, along with their sending times, are documented. Daily interviews were audio-recorded and transcribed. The recorded data and transcriptions were independently reviewed by three researchers, who coded them based on two main themes: types of smartphone use for question (2) and factors influencing persuasion effectiveness for questions (1), (3), and (4). Subsequently, the researchers convened to discuss the codes until a consensus was reached. Following that, a thorough review of all transcriptions was conducted to ensure the accuracy of the coding.}

\subsubsection{Semi-structured Interview Study}
\label{subsub:semi-interview}
\review{Our WoZ study provided insights into participants' problematic use behavior and reactions towards persuasion. To obtain a deeper understanding of the user's mental states during phone use,} we conducted a semi-structured interview study with another participant group.
We asked participants to recollect instances of problematic smartphone use. \review{Our interview started with the question: \textit{``When would you want to use an intervention app to limit your smartphone use?''} We then sought details about the scenario (\eg time, place, and concurrent activities) and user behaviors and reactions (\eg usage duration, feelings, and reflections).
Next, we asked participants to share their mental states during those instances. We asked questions: \textit{``Why do you use your phone even though you think you should not? What's your mental state behind these reasons?''}}  We followed the participants' lead during the interview.

All interviews were audio recorded and transcribed. \review{
Three researchers independently examined the transcriptions and coded the mental states in different smartphone use cases and contexts. Then they met and discussed the codes until reaching a consensus. To ensure coding accuracy, they went through all transcriptions one more time.
}

\subsection{Main Takeaways about Problematic Smartphone Use} 
\label{sub:formative_studies:mental_state}

\review{We summarized our main takeaways from the WoZ study and the interview study below. Table~\ref{tab:takeaways} in the Appendix summarizes our findings with participants' quotes. Table~\ref{tab:mental states with activity} summarizes the findings about mental states in Takeaway \textcircled{\raisebox{-.9pt} {3}} and Takeaway \textcircled{\raisebox{-.9pt} {4}}}.

\begin{table*}[]
\centering
\small
\begin{tabular}{p{0.45\textwidth}p{0.45\textwidth}}
\toprule
\multicolumn{2}{c}{\textbf{Boredom}} \\
\midrule
\multicolumn{1}{c}{\textbf{\textit{Engaging in Activities}}} & \multicolumn{1}{c}{\textbf{\textit{Not Engaging in Activities}}} \\
Users find current tasks boring, lack interest, and struggle to concentrate. This could be due to the task lacking challenges, not being sufficiently engaging, having high repetition, and not aligning with the user's genuine interests and desires. &
  Users feel bored with daily living in general, lack passion, and have no enthusiasm for engaging in activities. This might be because they have nothing to do, don't know how to pass the time, lack excitement in life, and feel that living is meaningless. \\
\hline
\multicolumn{2}{c}{\textbf{Stress}} \\
\hline
\multicolumn{1}{c}{\textbf{\textit{Engaging in Activities}}} & \multicolumn{1}{c}{\textbf{\textit{Not Engaging in Activities}}} \\
Users feel stressed about current tasks due to challenges posed by the environment, which deplete their resources and result in feelings of tension and unhappiness. This might be due to the abundance, difficulty, and urgency of tasks, causing users to feel anxious, fatigued, and lacking confidence in their abilities, leading to a pessimistic view of the outcomes. &
  Users feel stressed in the face of daily living and challenges from the environment that deplete their resources, making them feel tense and unhappy. This might be due to setbacks and unexpected events in life that users struggle to adapt to, leading to a pessimistic outlook on the future. \\
\hline
\multicolumn{2}{c}{\textbf{Inertia}} \\
\hline
\multicolumn{1}{c}{\textbf{\textit{Engaging in Activities}}} & \multicolumn{1}{c}{\textbf{\textit{Not Engaging in Activities}}} \\
Users find it difficult to transition from their current state to start the next activity, but without explicit negative emotions. This might be due to procrastination has become a habit, and there's insufficient motivation for the next activity. &
  Users indulge in idle inertia, but without specific negative emotions. This might be due to idling around has become a habit, and there's no motivation to organize new activities. \\
\hline
\multicolumn{2}{c}{\textbf{Others}} \\
\multicolumn{2}{c}{Beyond the scope of the current paper.} \\
\bottomrule
\end{tabular}
\caption{\review{Summary of Users' Mental States behind Habitual Smartphone Use.}}
\label{tab:my-table}
\label{tab:mental states with activity}
\Description{The table is organized into seven rows and two columns, including headers. The first, third, and fifth rows combine the two columns with the headers "Boredom", "Stress", and "Inertia" respectively. In the second, fourth, and sixth rows, the left column contains concrete descriptions of boredom, stress, and inertia under the condition of "engaging in activities"; the right column contains concrete descriptions of boredom, stress, and inertia under the condition of "not engaging in activities". The seventh row combines the two columns with the header "Others", referring to mental states "beyond the scope of the current paper". }
\end{table*}

\review{\textbf{Takeaway \textcircled{\raisebox{-.9pt} {1}}}} \textbf{Interventions for problematic smartphone use should target habitual usage.} Our WoZ study delineated two primary types of smartphone usage: instrumental and habitual, consistent with prior research categorization ~\cite{lukoff2018makes}.
We found that only habitual smartphone use warrants intervention. Interventions during instrumental use often led to user dissatisfaction. Notably, relaxation emerged as a crucial form of instrumental use, where participants deliberately used their phones to unwind or reward themselves after intense work or study. Intervening at such times was considered intrusive and inappropriate. The finding aligns with the Dual Systems Theory. In smartphone interactions, instrumental use relies on conscious decision-making (System 2), while habitual use is more instinctive (System 1). Hence, interventions should primarily target habitual use, which is also supported by earlier studies ~\cite{lukoff2018makes}.

\review{\textbf{Takeaway \textcircled{\raisebox{-.9pt} {2}}}} \textbf{The effectiveness of interventions depends on the alignment with users' mental states, personal goals, and contextual information}.
This is consistent with the literature, suggesting that a shift away from System 2 is due to emotional fluctuations and the absence of defined goals and intentions~\cite{lyngs2019self}. We experimented with different persuasive message content during our WoZ study. We found that when we incorporated users' mental states as a factor, which was inferred by their physical contexts and app usage patterns, into generating persuasive message content, participants were more willing to accept the intervention. Furthermore, highlighting users' personal goals enhanced intervention effectiveness. For instance, sending messages like, ``\textit{When you find yourself with idle time, consider engaging in meaningful activities such as reading, writing, or drawing}'' proved effective when users were in an idle state and had a goal for self-improvement. Our findings are supported by prior studies linking habitual smartphone use to specific mental states ~\cite{anderson2021habits,bayer2018technology}.

\review{\textbf{Takeaway \textcircled{\raisebox{-.9pt} {3}}}} \textbf{Semi-structured interviews revealed three primary mental states connected to habitual smartphone use: \textbf{\textit{boredom}, \textit{stress}, }and\textbf{ \textit{inertia}}.}
\begin{s_itemize}
    \item \textbf{Boredom} is an affective state characterized by low arousal and dissatisfaction due to insufficient stimulation ~\cite{fisherl1993boredom,mikulas1993essence}. The WoZ and interview studies identified common scenarios leading to boredom: (1) when the task at hand is too simple, lacking a balance between skill and challenge, \review{such as \textit{"doing simple assignments light on cognitive engagement"} (S3)\footnote{This is the serial number of the participant in the semi-structured interview study.} ,} (2) lack of interest in the current activity, \review{such as \textit{"completing assignments is to relieve a burden, instead of reaching achievement"} (S6),} and (3) devoid of any engaging activities during idle moments, \review {such as \textit{"after returning to home"} when is \textit{"not yet time to sleep"} (S8)}. 
    \item \textbf{Stress} refers to cognitive and behavioral reactions to unpredictable and unmanageable stimuli~\cite{koolhaas2011stress}. Participants frequently use smartphones because they experience (1) heightened anxiety when work demands exceed their abilities, \review{such as \textit{"having a challenging bug to locate when doing programming assignment"} (S3) or \textit{"work not progressing well"} (S9)} and (2) uncertainty about whether something would have a positive outcome, \review{such as \textit{"not sure if can get a job offer"} (S10)}. This aligns with increasing evidence that links smartphone use to perceived stress~\cite{chiu2014relationship,samaha2016relationships,vahedi2018association}.
    \item \textbf{Inertia}, in our context, refers to a psychological resistance that makes users reluctant to change their current activity state. It is similar to the idea illustrated by literature such as emotional inertia~\cite{kuppens2010emotional} or decision-making inertia~\cite{alos2016inertia}. Participants stated they commonly used their phones habitually to avoid changing into a new activity state from an idle state, \review{\textit{"checking the phone before starting to do assignment"} (S1) or\textit{ "shifting from a relaxed state to a focused state"} (S2).} Unlike stress or boredom, inertia does not elicit overt negative emotions but impedes the shift to the next task. 
\end{s_itemize}

\review{\textbf{Takeaway \textcircled{\raisebox{-.9pt} {4}}}} \textbf{Engaging vs. Not Engaging in Activities (Table \ref{tab:mental states with activity}).}
\review{
We further noticed two nuanced categories within mental states.
The first category (``engaging in activities'') denotes that users have activities to complete while habitually using smartphones. Participants either got distracted from the ongoing boring or stressful activities (\eg \textit{"I find myself instinctively reaching for my phone in search of mental stimulation when doing simple assignments light on cognitive engagement"} (S3)) or procrastinated to face the upcoming activities (\eg \textit{"I was reluctant to start handling this challenging work that I scrolled my phone screen anxiously"} (S2)). In contrast, the second category (``not engaging in activities'') means users have no schedule or don't know what to do while using smartphones habitually (\eg \textit{"After getting off work and returning home, I collapse on the sofa and binge-watch Tiktok for one to two hours"} (S9)). Differentiating activity engagement states and combining with the three identified mental states lead to six granular categories of habitual smartphone use, making the persuasion strategies design more situated to users' scenarios. For users engaging in activities, the persuasion not only aims to stop them from smartphone use but also to encourage them to either continue or initiate their activities.}

\subsection{Persuasion Strategies Design}
\label{sub:formative_studies:strategy}

Based on the takeaways and inspired by the Dual Systems Theory and the Existence, Relatedness, and Growth (ERG) theory, we proposed four distinct persuasion strategies: \textbf{Understanding, Comforting, Evoking}, and \textbf{Scaffolding Habits}. Developing from Maslow’s Hierarchy of Needs, the ERG theory further summarizes human motivation into three levels: (i) the physiological and safety basic needs for existence, (ii) the social needs for feeling related and accepted, and (iii) the need to grow and self-actualize. The theory has been applied to workplaces to increase productivity and job satisfaction~\cite{arnolds2002compensation,caulton2012development}. 

\review{
Some of our takeaways align with these theories. For example, 
\textbf{Takeaway \textcircled{\raisebox{-.9pt} {1}}} aligns with the Dual Systems Theory and suggests that to avoid habitual smartphone use out of instinctive System 1, it is necessary to cultivate enough motivation to maintain conscious but difficult System 2. Moreover, \textbf{Takeaway \textcircled{\raisebox{-.9pt} {2}}} shows that persuasive messages relieving mental states and reminding personal goals are effective, which is consistent with the human motivation of relatedness and growth outlined in the ERG theory.}

\begin{figure}[h]
    \centering
    \includegraphics[width=1\linewidth]{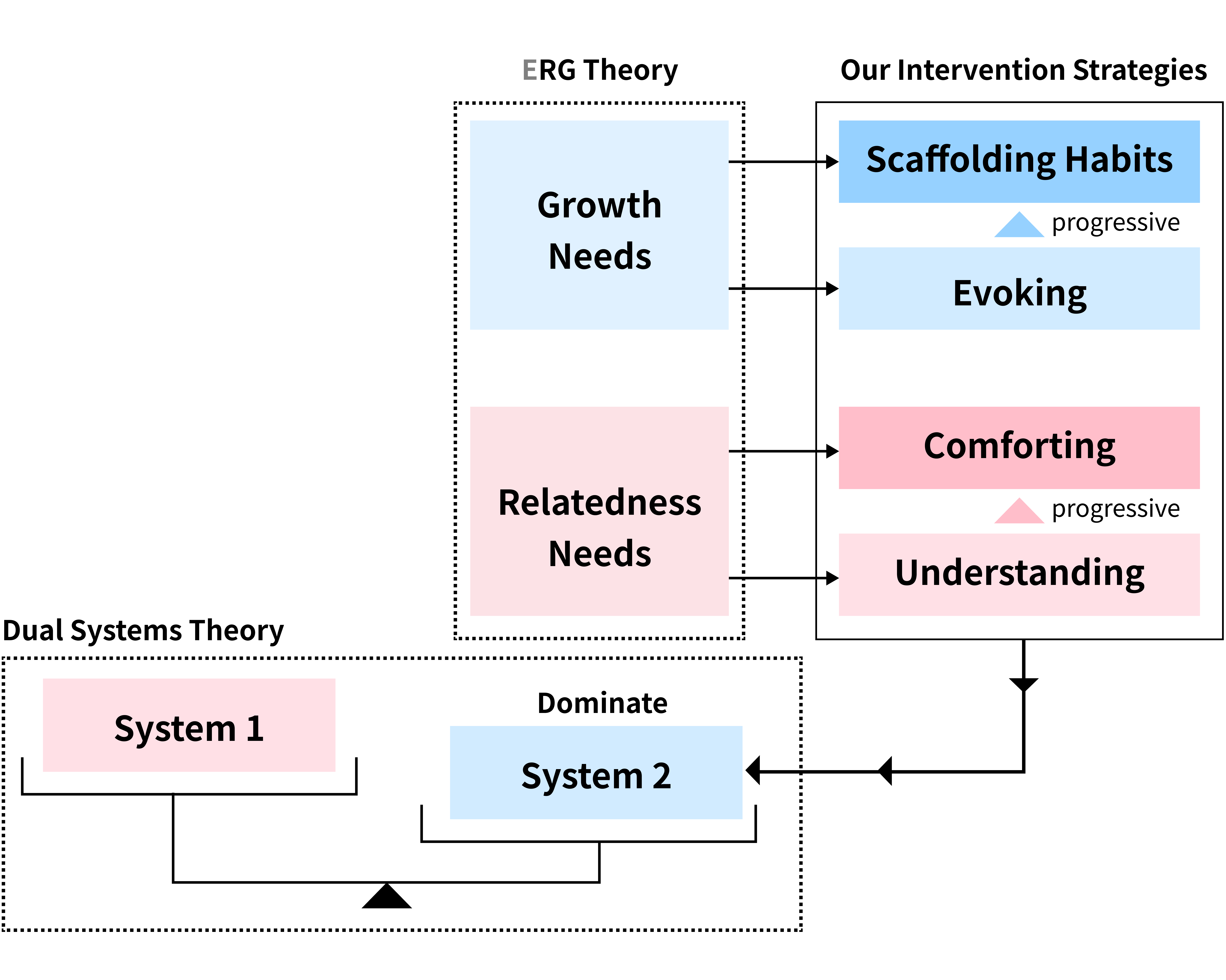}
    \caption{Mapping of Persuasion Strategies to the ERG Theory (Existence, Relatedness, and Growth) and the Dual Systems Theory. According to the Dual Systems Theory, there is a competitive relationship between System 1 (habitual smartphone use) and System 2 (meaningful activity), much like being placed on a scale. To make System 2 heavier than System 1, weights are added—Relatedness needs and Growth needs (the second and third levels of ERG Theory). To support users' Relatedness needs, we design two persuasion strategies: Understanding and Comforting. To motivate Growth needs, we design two other persuasion strategies: Evoking and Scaffolding Habits.}
    \Description{This is a picture of a balance scale. On the left side of the scale, there is a square labeled "System 1," and on the right side, there is a square labeled "System 2." These two squares are enclosed by dashed lines and labeled "Dual Systems Theory," indicating that we are applying the Dual Systems Theory. On the square labeled "System 2," there are also squares stacked in sequence with labels "Relatedness Needs" and "Growth Needs," and these two squares are enclosed by dashed lines and labeled "ERG Theory," where the letter "E" is colored gray, lighter than the other black letters, indicating that we are only using the second layer (relatedness) and the third layer (growth) of the ERG Theory. The right side of the scale sinks lower than the left side, and the square on System 2 is labeled "Dominate." This is meant to convey that after using the ERG Theory to stimulate user Related Needs and Growth Needs, System 2 can gain an advantage in competition with System 1. The square labeled "Related Needs" has arrows pointing to two squares labeled "Understanding" and "Comforting" respectively. The square labeled "Understanding" has an arrow pointing to the square labeled "Comforting," indicating that the Comforting intervention strategy goes a step further than the Understanding intervention strategy. The square labeled "Growth Needs" points to two squares labeled "Evoking" and "Scaffolding Habits" respectively. The square labeled "Scaffolding Habits" has an arrow pointing to the square labeled "Evoking," indicating that the Scaffolding Habits intervention strategy goes a step further than the Evoking intervention strategy.}
    \label{fig:mapping_of_strategy_to_erg}
\end{figure}

\begin{figure*}[h]
    \centering
    \includegraphics[width=0.95\linewidth]{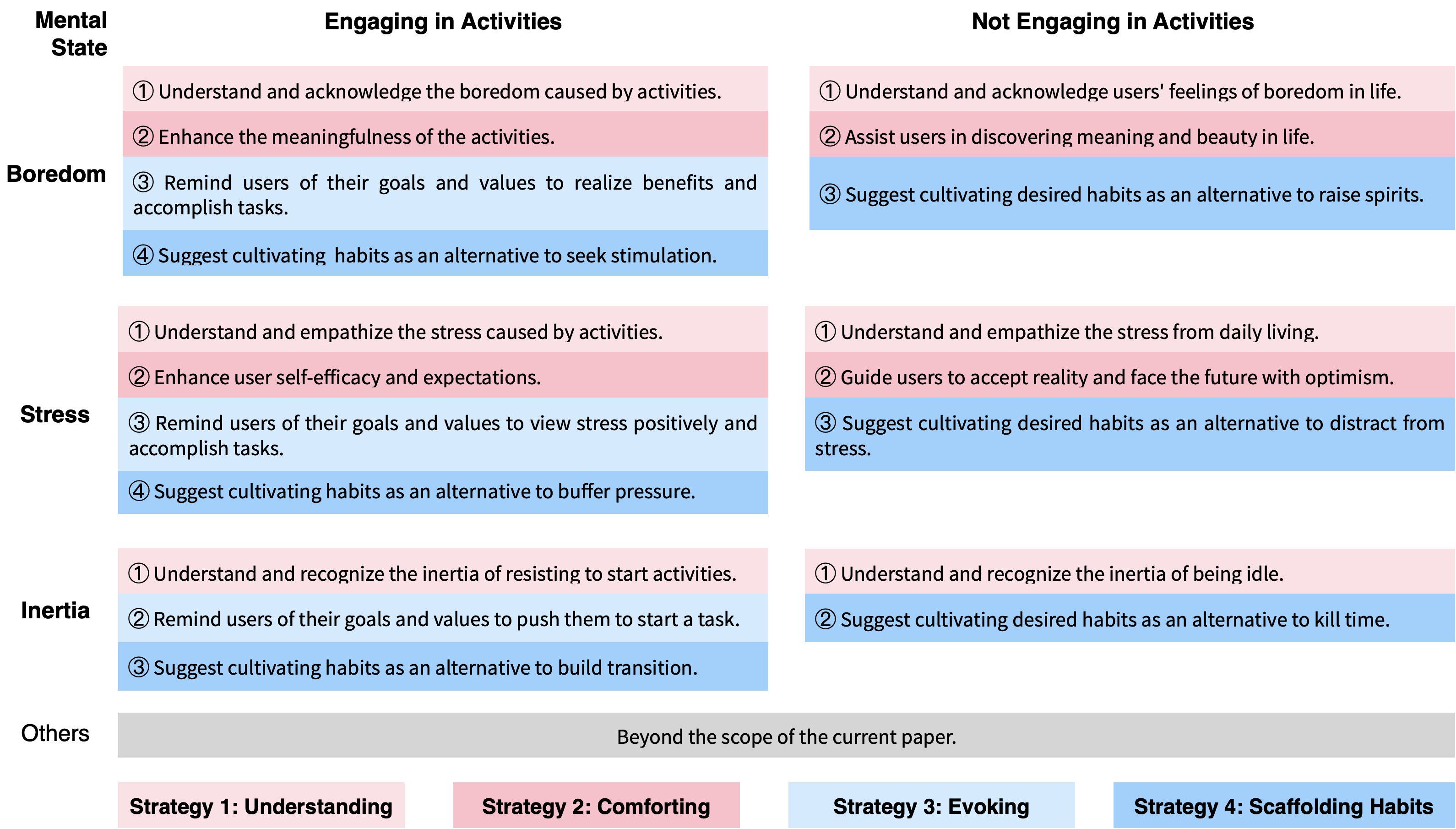}
    \caption{\review{Summary of Persuasion Strategies under Different Mental States. In the table, each color denotes the corresponding persuasion strategy applicable to this scenario.}}
    \label{fig:strategy}
    \Description{This is a 3 x 2 table, with three mental states (boredom, stress, inertia) and two activity states (engaging in activities, not engaging in activities). After the "three mental states" column, the fourth column is an "Others" column, referring to mental states "beyond the scope of the current paper". At the bottom of the table, there are four squares labeled "Understanding," "Comforting," "Evoking," and "Scaffolding Habits." In the "boredom × engaging in activities" cell, there are four rows of concrete descriptions of "Understanding," "Comforting," "Evoking," and "Scaffolding Habits." In the "boredom × not engaging in activities" cell, there are three rows of concrete descriptions of "Understanding," "Comforting," and "Scaffolding Habits." In the "stress × engaging in activities" cell, there are four rows of concrete descriptions of "Understanding," "Comforting," "Evoking," and "Scaffolding Habits." In the "stress × not engaging in activities" cell, there are three rows of concrete descripyions of "Understanding," "Comforting," and "Scaffolding Habits." In the "inertia × engaging in activities" cell, there are three rows of concrete descriptions of "Understanding," "Evoking," and "Scaffolding Habits." In the "inertia × not engaging in activities" cell, there are two rows of concrete descriptions of "Understanding," and "Scaffolding Habits."}
\end{figure*}

Accordingly, we map strategies to the relatedness level and growth level of the ERG theory to arouse users' motivation for System 2 (Figure~\ref{fig:mapping_of_strategy_to_erg}). The existence level concerning physiological and safety needs is not included in our theoretical framework. At the level of relatedness, \textbf{Understanding} and \textbf{Comforting} aim to empathize with users' emotions, offering support and empowering users to manage System 2. At the growth level, \textbf{Evoking} reminds users of their personal development goals, and \textbf{Scaffolding Habits} guides them in replacing habitual smartphone use with activities conducive to self-fulfillment, thereby turning awareness into action. Then, we map 4 strategies to 3×2 mental states in Figure \ref{fig:strategy}.

\subsubsection{Understanding} 
\textit{Understanding} is a critical strategy to motivate users at the Relatedness level. Past literature suggests that seeking understanding is a coping mechanism~\cite{compas2001coping}, and chatbots' empathetic expressions are favored over emotionally neutral advice~\cite{liu2018should}. Therefore, \textit{Understanding} covers all mental states. This example shows how we integrate understanding into the persuasive content intervention: \textit{"Hi, I know that sometimes you may feel bored and lacking interest. It's okay, this is a very normal feeling. Everyone goes through such times."}

\subsubsection{Comforting} 
\textit{Comforting} aims to comfort users who are experiencing emotional fluctuations, which cause them to shift away from System 2.  Prior studies suggest that coping with boredom should focus on meaningfulness ~\cite{nett2010feeling}. For example,  we design a persuasive message to say \textit{"Hey, I know some things might seem a bit boring, but sometimes we need to find the fun in them. Have you ever thought that completing this task would bring you closer to your goals?"} In addition, encouragement, humor, acceptance, and wishful thinking~\cite{compas2001coping} can be used to cope with stress by lightening the uncontrollable and unpredictable nature of stressors. For example, \textit{"Don't worry, you have the capability to complete the task! Believe in yourself, and the outcome will pleasantly surprise you."} In the state of inertia without explicit negative emotions, \textit{Comforting} is not employed.

\subsubsection{Evoking} 
Evoking personal goals is a compelling, persuasive technique based on the WoZ study. Literature suggests that goals and values are important for people to sustain System 2~\cite{lyngs2019self} and are closely related to growth motivation. Thus, \textit{Evoking} considers users' goals (\eg getting high scores in exams, achieving academic success) for designing persuasive strategies. For example, \textit{"Hi! I know you want to play with your phone, but completing tasks is crucial for your IELTS! Keep going, and you're one step closer to a high score!"} This strategy is applied only to scenarios where users are engaging in activities. Goals are arguments used to encourage them to complete or initiate their tasks. In scenarios where users are not engaging in activities, \textit{Scaffolding Habits} assumes the function of \textit{Evoking} by recommending activities that correspond with users' goals, as stated below.

\subsubsection{Scaffolding Habits}
Last, we encourage users to develop alternative beneficial habits to habitual smartphone use, aligning with their personal value and growth need~\cite{lyngs2019self,pinder2018digital}. By pre-identifying users' preferred habits and considering variables like location and time of habitual smartphone use, we suggest appropriate substitutes to assist users in \textit{Scaffolding Habits}. For example, \textit{"Hi, why not use this moment to memorize vocabulary instead of using your phone? It can help you learn a language and achieve your goals faster!"} \textit{Scaffolding Habits} covers all mental states.

\section{MindShift DESIGN}
\label{sec:design}

Building on top of the persuasion strategies we propose in the previous section, we introduce the design of our intervention system: generating persuasive content (what content to intervene with, Sec. ~\ref{sub:design:what}), interaction flow (how to intervene, Sec. ~\ref{sub:design:how}), and intervention timing (when to intervene, Sec. ~\ref{sub:design:when}).

\subsection{What Content to Intervene with: LLM-Powered Persuasive Content Generation}
\label{sub:design:what}
\review{We first delved into the importance of context and mental states in persuasion content generation (Sec.~\ref{subsub:context_and_strategy}). After establishing the significance of context and mental states, we explored how these elements can be intricately integrated into the prompt design (Sec.~\ref{subsub:prompt_design}).}

\subsubsection{\review{Context and Mental States in Generating Persuasive Content}}
\label{subsub:context_and_strategy}
\review{As suggested in the \textbf{Takeaway \textcircled{\raisebox{-.9pt} {2}}} in Sec.~\ref{sub:formative_studies:mental_state}, the effectiveness of interventions also depends on contextual information.
We presented a test case with examples to demonstrate the impact of context and mental states on content generation.
In this case, we first constructed a typical college student's context, including time (at late night 00:36 AM), location (in the dorm), and phone usage data (5 mins since the last habitual usage and 10 mins current habitual usage).
We then outlined the user's assigned mental state (stressed - engaging in activities),  the user's goals (growing research skills, staying healthy) and habits (enjoying outdoor activities), and the corresponding four strategies based on Figure~\ref{fig:strategy}.
}
\begin{table*}[t]
\caption{\review{Examples of Generating Persuasive Content with A Test Case Study on Different Context and Mental State Combinations. The letters in parentheses at the end of the sentences indicate the strategies used during sentence generation: U for Understanding, C for Comforting, E for Evoking, S for Scaffolding Habits, and N for No Strategy}}
\label{tab:prompt-comparsion}
\Description{The table is organized into three rows and three columns, including headers. The first column categorizes the content generation approach into two types from up to bottom: "With Context" and "Without Context."  The first row is the primary header, clarifying the mental state consideration: "With Mental State" and "Without Mental State." from left to right. Each combination of context and mental state presence or absence in the table aims to showcase different approaches to persuasive content generation. In row two, under the "With Context" category, the table details examples of generated sentences considering or disregarding the user's mental state. In rows three, under the "Without Context" category, present sentences that either address or ignore the user's mental state. If the sentence is "With Mental State", it will use strategies like understanding, comforting, and evoking, indicated by letters (U), (C), (E), and (S). }
\small 
\begin{tabular}{>{\centering\arraybackslash}m{0.5cm}m{6cm}m{6cm}}
\hline
\hline
 & \multicolumn{1}{c}{\textbf{With Mental State}} & \multicolumn{1}{c}{\textbf{Without Mental State}} \\ 
\hline
\multicolumn{1}{c}{\textbf{With Context}} & 
1. It's already 00:36, staying up late is stressful, you're not alone, we all feel this way.\textit{ (U)} \newline
2. Hey, you've got this! Embrace the challenge with a smile. You're unstoppable, even at 00:36 AM! \textit{ (C)} \newline
3. Think about that research article you're about to complete, turn the stress into motivation!\textit{ (E)} \newline
4. Try putting down your phone, enjoy the night sky outside the window, relax your eyes.\textit{ (S)} & 
1. It's already 00:36, your phone should probably rest too. \textit{ (N)} \newline
2. At this time in the dorm, how about closing your eyes and resting? \textit{ (N)} \newline
3. You've already enjoyed 10 minutes of the digital world. \textit{ (N)} \newline
4. Just 5 minutes ago, we just said goodbye, meeting again? \textit{ (N)} \\
\hline
\multicolumn{1}{c}{\textbf{Without Context}} & 
1. Understand your anxiety, it's normal, accept your emotions.\textit{ (U)} \newline
2. Relax, each task is a stepping stone to achievement, be optimistic.\textit{ (C)} \newline
3. Keep going, you will be closer to completing that great research job.\textit{ (E)} \newline
4. How about relaxing in a different way, look far away, let your eyes rest too.\textit{ (S)} & 
1. Friend, your phone might need a rest, and so do your eyes! \textit{ (N)} \newline
2. Try putting down your phone, and take a look at the outdoor night views. \textit{ (N)}\newline
3. The phone in your hand is not the world, the real fun is around you! \textit{ (N)} \newline
4. Every time you put down your phone, it's an opportunity to add points to life! \textit{ (N)} \\
\hline
\bottomrule
\end{tabular}
\end{table*}

\review{Differing in context and mental state inclusion, we used GPT-3.5 to generate four sets of persuasive content examples.}
\review{Table~\ref{tab:prompt-comparsion} lists the examples of GPT-3.5's outputs, demonstrating  that both contextual information and mental state guidance improve content quality. 
Contextual data empowers GPT to tailor its outputs to the user's current situation, including more poetic sentence phrases such as \textit{"enjoy the night sky outside the window"} or contextual information such as \textit{"You're unstoppable, even at 00:36 AM"}. Adding the mental state guidance enhances GPT's ability to assist users in managing their negative emotions, including empathetic messages such as \textit{"you are not alone"} and encouraging phrases such as \textit{"be closer to completing that great research job"}.}

\subsubsection{Prompt Design}
\label{subsub:prompt_design}
\begin{figure*}[ht]
    \centering
    \includegraphics[width=0.99\textwidth]{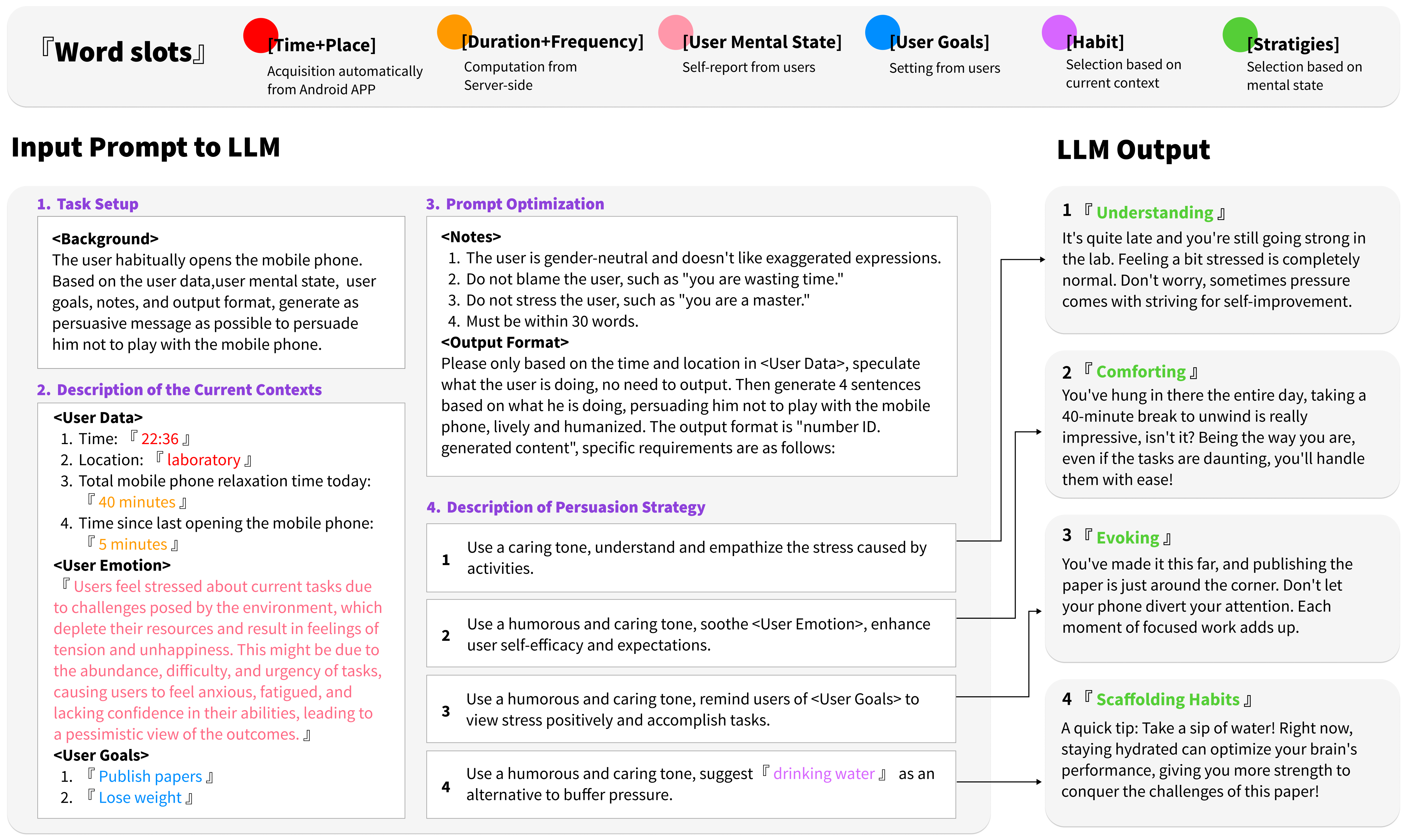}
    \caption{Prompt Templates Used to Generate Persuasive Content with GPT-3.5. The word slots (in the top box) represent different categories of information to be filled in based on the user's current situation. The color indicates the mapping between the slots and the input prompt.
    The small words under each slot explain the source of the content. 
    The input prompt consists of four parts: (1) Task Setup; (2) Description of the Current Contexts; (3) Prompt Optimization (to improve language quality and reduce harmful content); and (4) Description of Persuasion Strategy (as introduced in Sec.~\ref{sec:formative_studies} and Figure~\ref{fig:strategy}).
    The LLM output shows the persuasion example of four strategies when the user's mental state is ``stressed, engaging in an activity''.
    } 
    \label{fig:prompt-design}
    \Description{The image is divided into three major sections: the top area labeled as "Word Slots", the bottom left corner labeled as "Input Prompt to LLM", and the bottom right corner labeled as "LLM Output".
In the "Word Slots" section at the top, there are six subsections. Each subsection consists of a colored circle for easy identification, a title indicating the category, and content that specifies the source of that category's information. From left to right, these are:
1. A red circle labeled "Time+Place", with the source noted as "Acquisition automatically from Android APP".
2. An orange circle labeled "Duration+Frequency", with the source being "Computation from Server-side".
3. A pink circle labeled "User Mental State", with the source being "Self-report from users".
4. A blue circle labeled "User's Goal", with the source being "Set by users".
5. A circle with no color specified labeled "Habits", with the source as "Selection based on current context".
6. A green circle labeled "Strategies", with the source being "Selection based on mental states".
In the bottom left corner, labeled as "Input Prompt to LLM", there are four text boxes arranged in two rows and two columns. From top to bottom and left to right, these boxes are:
1. "Task Setup," which contains a description of the task background.
2. "Description of the Current Contexts," which includes three modules: "User Data," "User Mental State," and "User Goals." Texts that need to be filled with word slots are highlighted in corresponding colors.
3. "Prompt Optimization," which contains two modules: "Notes" and "Output Format".
4. "Description of Persuasion Strategy," which describes four different strategies—Understanding, Comforting, Evoking, and Scaffolding Habit—in that order. Arrows point from each strategy to its corresponding content in the "LLM Output" section.
Finally, in the bottom right corner, labeled as "LLM Output", there are four text boxes. From top to bottom are Understanding, Comforting, Evoking, and Scaffolding Habits, each containing corresponding persuasive messages.}
\end{figure*}

We constructed four important prompt input factors and fed them to GPT to generate high-quality and persuasive content. As illustrated by purple text in Figure \ref{fig:prompt-design}, four factors are arranged in a sequence from 1 to 4. We then concatenated them to build a complete prompt as the input to LLMs to generate persuasive content. We tested both GPT-3.5 and GPT-4 and chose to adopt GPT-3.5 as our target LLM to strike a balance between the content generation quality and the speed\footnote{GPT-4 introduces a long lag and negatively impacts user experience. Moreover, as we introduce below, the prompts we used as the input for GPT-3.5 include general information that is not individually identifiable. However, we do acknowledge the privacy risk of our method. We will have more discussion in Sec.~\ref{sec:discussion}}. 

We lay out the details about how we designed the prompt below: 

(1) \textbf{Task Setup}:
As shown in the left top box in the ``Input Prompt to LLM'' in Figure \ref{fig:prompt-design},  Task setup includes <Background> module, providing GPT-3.5 with the global instructions.

(2) \textbf{Description of the Current Contexts}: 
To make each generated content contextually relevant and personalized, it is necessary to describe the user's current contexts (see the left bottom box in the ``Input Prompt to LLM'' in Figure \ref{fig:prompt-design}). This part includes three modules:
(a) The <User Data> module describes the user's real-time physical context, including the current time, location, habitual phone usage duration, and the time elapsed since the last habitual phone check.  This is collected through phone sensors, see more details in Sec.~\ref{sub:implementation:client}.
(b) The <User Mental State> module includes users' input from their devices regarding negative emotions and activities. The prompt for this module is selected from the mental state definition (Table \ref{tab:mental states with activity}). This is collected through real-time self-report, see Sec.~\ref{sub:design:how}. 
(c) Finally, the <User Goals> module describes what the user values and plays a crucial role in the generation of\textit{ Evoking} strategies. Specific user goals are collected during the initialization  (Step 0 in Figure \ref{fig:interaction-process}). This is collected through the initial setup, see Sec.~\ref{sub:design:how}.
The elements mentioned above are represented as word slots (enclosed in brackets in the input prompt), where the users' actual context information can be inserted.

(3) \textbf{Prompt Optimization}:
To improve the  GPT's content quality and effectiveness, we carefully crafted the prompt according to OpenAI's official guidelines \cite{OpenAI_best_practice}.
Sometimes LLMs can generate harmful, offensive, or biased texts \cite{zhuo2023exploring,hartvigsen2022toxigen}. We employed an iterative prompt design process to ensure that the persuasive content is appropriate. 
Initially, one researcher created initial prompts and generated content using GPT-3.5, \review{for six mental states (Figure \ref{fig:strategy}) with five iterations each.} Subsequently, two other researchers rated the satisfaction level of the generated content (scoring from 1 to 5), iterating until average satisfaction exceeded 4 to create content that is concise, appropriate, and engaging, while also aligned with our persuasive strategy. 
\review{Through this process, we addressed some issues with LLM-generated content, such as gender-biased expressions and deviations from human preferences (like being overly exaggerated or stressful) by adding an additional <Notes> section to instruct LLMs' generation.
We also took steps to prevent hallucinations by avoiding certain real-world fact-related prompt statements that LLMs can easily make mistakes.
By specifying the need to consider current activities in the <output format>, we've made the outputs more contextually relevant.} 
We note that this process cannot fully address the ethical concerns, which we further discuss in Sec.~\ref{subsub:LLMRisk}.
Our final optimization prompt, shown in the middle top box in Figure \ref{fig:prompt-design}, includes two modules:
(a) The <Notes> module clarifies restrictions;  (b) The <Output Format> module guarantees the correct output format.

(4) \textbf{Description of Persuasion Strategy}:
To ensure each generated content aligned with our proposed strategies, we provide a short description of each strategy based on our design in Sec.~\ref{sec:formative_studies} and Figure~\ref{fig:strategy}, as indicated by the middle bottom box in Figure \ref{fig:prompt-design}. Based on users' in-the-moment mental states, we will select the corresponding strategies. Note that for the \textit{scaffolding habits} strategy, we also input a habit selected from users' initialization.

\subsection{How to Intervene: Interaction Flow}
\label{sub:design:how}
\begin{figure*}[]
    \centering
    \includegraphics[width=\textwidth]{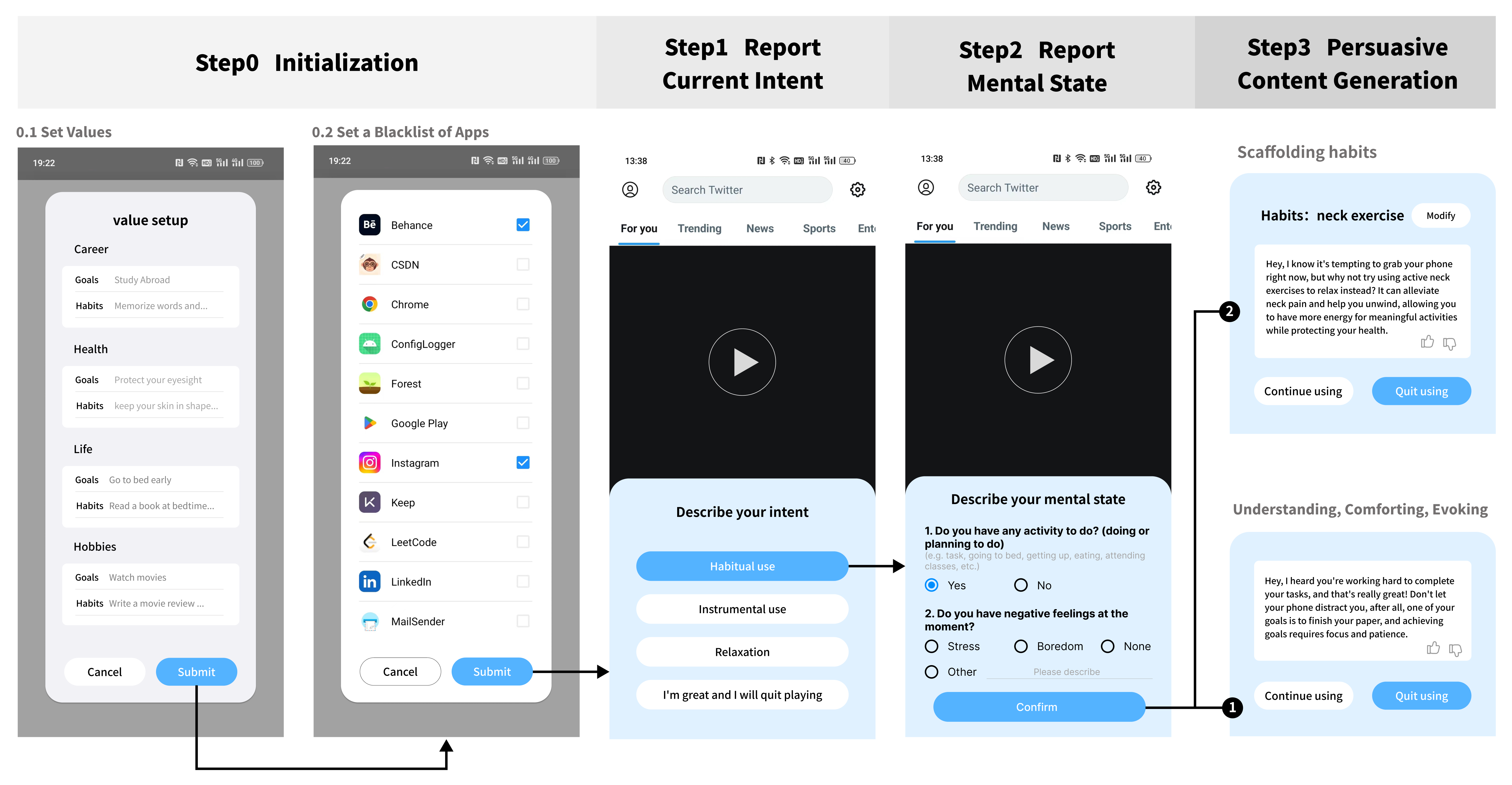}
    \caption{\minorreview{Interaction Flow.} Users first complete the global settings for their value and app list (Step 0). When opening a black-listed app, users need to first self-report their phone usage intent (Step 1). If the intent is habitual use, the app asks them to report their current mental state (Step 2). After that, a corresponding persuasion shows (Step 3).}
    \label{fig:interaction-process}
    \Description{Three figures about the intervention interface design. (Left) Two initialization pages. The left is about value setup, where there is a list of four value items and each item has two text edit boxes: goals, and habits. The value items are Career, Health, Life, and Hobbies. The right is about the blacklist of apps setup, where there is a list of apps installed on users' phones, each accompanied by a checkbox.
(Middle Left) The intent report intervention pop-up interface, and it has four buttons to choose from: habitual use, instrumental use, relax, I'm great and I will quit playing. (Middle Right) The mental state report intervention pop-up interface, and it has two single-choice questions: (1) Do you have any activity to do("Yes" or "No"), and (2) Do you have negative feelings at the moment, including options of "Stress", "Boredom", "None", or "Other Negative Feelings". The "Other Negative Feelings" option has an edit text box next to it for description. (Right) Two persuasion pop-up interfaces. The top half is an interface for scaffolding habits and persuasion strategy. The interface is divided into three main sections. At the top, a row presents recommended habits,  with an adjacent 'modify' button for customization. In the center, a rectangular text box displays auto-generated persuasive content. The bottom right corner of this text box features 'thumb-up' and 'thumb-down' buttons for user feedback. At the bottom, there are two buttons labeled 'continue using' and 'quit using,' offering options to proceed or discontinue. The bottom half is an interface for understanding, comforting, and evoking strategy. This interface is similar to the previous one but lacks the top row of recommended habits. It still features the central text box with autogenerated persuasive content and 'thumb-up' and 'thumb-down' buttons at the bottom right. The options 'continue using' and 'quit using' remain at the bottom.}
\end{figure*}

\review{This section outlines the design of the interaction flow in our application.}
\review{Our interaction process needs to achieve three functions: (1) collect users' phone usage intent to identify habitual use, (2) obtain necessary information for  prompt construction, and (3) display the generated content.}

\review{Following Takeaway \textcircled{\raisebox{-.9pt} {1}}  in Sec.~\ref{sub:formative_studies:mental_state}, interventions should be targeted at habitual usage. Since automatic detection methods are unreliable (discussed further in Sec.~\ref{sec:discussion}), we ask users to self-report, and the system only triggers intervention when the user reports habitual use.}
\review{In our prompt design in Sec.~\ref{subsub:prompt_design}, we need two categories of information from users: their goals and habits, and their mental state.
Goals and habits tend to be stable. So we integrated them into the app's settings page, and users could adjust them as needed.
Mental states, however, are more dynamic. Therefore, we captured them through participants' self-reporting.
In summary, each intervention episode includes three steps: (Step 1) the user reports their usage intent, (Step 2) the mental state, and (Step 3) the corresponding generated content is displayed, as shown in Fig.~\ref{fig:interaction-process}. Our final design is as follows:}

\textbf{Step 0: Initialization}. When initiating the MindShift app, we ask users to complete two global settings. The first is to set their values in four categories: career, health, life, and hobbies, detailing their goals and habits in each. This process serves two purposes: first, it provides the necessary goals for our \textit{Evoking} strategy; second, it enables the creation of personalized habits in the \textit{Scaffolding Habits} strategy. The second is to set a blacklist of apps. Launching apps from this list will trigger intervention.

\textbf{Step 1: Intent report}. Interventions are only necessary when users habitually use their phones (\textbf{Takeaway \textcircled{\raisebox{-.9pt} {1}}}).
To collect users' intents, we employ a self-reporting approach. Every time \review{a blacklisted app is first opened during an unlock session}, users choose from three options: "Habitual use", "Instrumental use", and "Relaxation". Only when users select "Habitual use", the intervention will proceed.
Automatically detecting all use is beyond the scope of this paper. We envision that future work can automate this process, as discussed in Sec.~\ref{sec:discussion}.

\textbf{Step 2: Current mental state report}. The mental state is a key factor that triggers users' habitual phone use. Unlike physical context, detecting mental state is challenging due to the lack of mature techniques. Therefore, we ask users to self-report. Based on the mental states listed in Sec.~\ref{sub:formative_studies:mental_state} and \textbf{Takeaway \textcircled{\raisebox{-.9pt} {3}} \& \textcircled{\raisebox{-.9pt} {4}}},  we propose two single-choice questions for users to report their mental state: (1) whether they are engaged in activities ("Yes" or "No"), and (2) whether they currently have any negative feelings, including options of "Stress", "Boredom", "None" (\ie inertia), or "Other Negative Feelings".
The "Other Negative Feelings" option, with an adjacent text box for specifics, covers unlisted emotions.
\review{We tested these questions for reliability and validity~\cite{price2015reliability}. To ensure validity (\ie accurate reflection of mental states), three psychology experts verified the alignment of our questions with mental state coding in Sec.~\ref{sec:formative_studies}. 
To ensure reliability (answer consistency), we asked three pilot study participants to respond to situational mental state descriptions with two single-choice questions, achieving uniform responses.\footnote{We plan to conduct more comprehensive validity and reliability testing in future work, as discussed in Sec.~\ref{sec:discussion}.}}

\textbf{Step 3: Persuasion}. Based on our intervention content design in Figure~\ref{fig:prompt-design}, we leverage the power of GPT-3.5 to generate persuasive messages.
The messages are displayed in a pop-up window, where users can choose to either quit the app or continue using it (the bottom of Step 3 in Figure~\ref{fig:interaction-process}). Additionally, users can also provide optional feedback by giving a thumbs up or down.

Furthermore, for the\textit{ Scaffolding Habits} strategy, it is essential to link users' own habits to specific use contexts. Therefore, we implement a user participation mechanism, adding an additional habit item along with an edit button (the upper interface of Step 3 in Figure \ref{fig:interaction-process}). The habit item represents a system-generated suggestion based on the user's current context and initial settings in Step 0. Users can edit and update their desired habits. Once submitted, the modified habit will be recommended the next time when users are in the same context.

\subsection{When to Intervene: Intervention Trigger Mechanism}
\label{sub:design:when}

\review{We consider the user's intent of use in the intervention trigger rule (Step 1 of Sec. \ref{sub:design:how}). Specifically, interventions will be triggered when a user's self-report intent is habitual use.}

As in Figure~\ref{fig:strategy}, each mental state allows for multiple persuasion strategies. We devise a simple procedure. After determining the mental state and narrowing down the specific strategies, we first randomly sample one strategy and generate an intervention message. Then, we loop over other strategies and show new strategy messages every two minutes until the users leave the app. After looping over all appropriate strategies under this mental state, the intervention will stop.
\review{The usage duration is calculated based on the total usage time of a single blacklisted app during one unlocking session.}
\review{The two-minute interval setting is derived from the statistical analysis in the WoZ study,  where 90\% of users spent less than 5 minutes on a blacklisted app in a single session.
We thus set the interval between interventions as two minutes \minorreview{as a convenient delay} to facilitate the exploration of different strategies, as further supported by the analysis in Sec.~\ref{subsub:habitual}.}

Users only need to report their habitual use and mental states once (\ie Step 1 and 2 in Figure~\ref{fig:interaction-process}) when they open a specific app during each screen unlock session. This design is based on three considerations: \review{(1) The initial mental state when opening an app is crucial, as it triggers habitual use (Sec.~\ref{sub:formative_studies:mental_state}). (2) As reflected in our pilot study in Sec.~\ref{sub:design:how}, multiple reports during app switching can be annoying and negatively affect user experience. (3) Some previous studies suggest that users' stress level tends to be retained even with coping techniques~\cite{duvenage2020technology}. We assume this also applies to other mental states, so most users' mental state remains stable during one session (90\% was less than 5 minutes), as further supported by the analysis in Sec. ~\ref{subsub:res-usage}.}
\review{We also discuss future potential ways to enhance accuracy in Sec. \ref{subsub:implication} \& \ref{sub:discussion:limitation}.}

\section{System Implementation}
\label{sec:implementation}

We built an Android application to instantiate our design. Our system consists of a client and a server.

\subsection{Client-Side Implementation}
\label{sub:implementation:client}

The client-side is an Android app that implements all the features of our design. We use accessibility services to detect the opening and closing of apps on the phone. 
The client is also responsible for collecting and uploading data to the server, including screen's off and unlock status, application name, application opening and closing times, and location data obtained through the Amap API~\cite{amap}.
To prevent the accidental killing of the accessibility service and ensure compliance, our app includes a background service checking the accessibility service status every 5 minutes. If it detects that the service is terminated, the app informs the server to email a researcher, who then reminds the user to reactivate the service, ensuring data integrity.

\subsection{Server-Side Implementation}

The server side is responsible for generating persuasive content through four key tasks, ensuring that the intervention is personalized, contextually relevant, and delivered in a timely manner.

(1) \textbf{User Data Computation}: The server processes user data from client-uploaded app data for use in word slots, including the phone's total habitual use time and last habitual opening time.

(2) \textbf{Habit Selection}: Next, the server selects a habit mostly matched with the current user's mental state, location (\ie the specific building), and time (\ie the hour of the day) from the users' initialization  (\ie Step 1 in Figure~\ref{fig:interaction-process}). To reinforce the habit-context link, unless users thumb down or modify it, the same habit will be recommended in the same context. More details can be seen in Step 4 of \ref{sub:design:how}.

(3) \textbf{Strategy Counterbalance}: To balance the frequency of each strategy across mental states, the server counterbalances strategy order in the prompt in Sec.~\ref{sub:design:how}.

(4) \textbf{Content Generation}:
After obtaining the user contexts, habits, and persuasion strategies, the server uses the OpenAI GPT-3.5 API to generate persuasive content. We adopted a streaming, character-by-character generation approach, allowing the persuasive content to start being displayed within 2 seconds.

\section{Field Experiment}
To evaluate the effectiveness of MindShift, we conducted a 5-week field experiment.
We introduce experimental design (Sec. \ref{subsub:baseline} \& \ref{subsub:exp-design}), participant recruitment (Sec. \ref{subsub:participant}), and experiment procedure (Sec. \ref{sub:exp-procedure}). 

\subsection{Baseline \review{and MindShift-Simple} Intervention Methods}
\label{subsub:baseline}
As \textit{MindShift} is one of the first persuasion intervention systems leveraging an LLM to generate dynamic persuasion content, there are no comparable systems other than the traditional persuasion techniques.
We compared \textit{MindShift} against a persuasive reminder baseline, one of the most commonly adopted intervention methods in commercial apps~\cite{Tiktok,apple}.
To ensure the fairness of the comparison, the baseline is designed to be the same as the intent report step, as illustrated in Step 1 in Figure~\ref{fig:interaction-process}. \review{Specifically, it only requires users to report the intent the first time they open the blacklist app after unlocking.
It can also be used to collect the proportion of users' initial intents, facilitating the analysis of the intervention effect of \textit{MindShift}.}
We name this intervention as \textit{Baseline}.

Moreover, in order to evaluate the effectiveness of the mental states and persuasion strategies proposed in Sec.~\ref{sec:formative_studies}, we further designed a simplified version, \textit{MindShift-Simple}, by removing the mental states and persuasion strategies from the prompt design. 
\review{Specifically, in the prompt design (Figure \ref{fig:prompt-design}), we retained only <User Data> in the (2) Description of the current context and removed the (4) Description of the persuasion strategy. Meanwhile, the last sentence in <Output Format> was changed to generate four sentences at once. We kept other setups consistent, ensuring that both versions' language features (such as both tone styles are humorous and caring) are as consistent as possible.}
\review{Examples of content generation in \textit{MindShift-Simple} are as shown in Table \ref{tab:prompt-comparsion} under `With Context' and `Without Strategy'.}

In total, we have three intervention methods to compare: \textit{Baseline}, \textit{MindShift-Simple}, and \textit{MindShift}. Table \ref{tab:intervention-type-description} shows the comparison. \review{ Figure \ref{fig:comparision-tech} in Appendix further shows their interaction flow.}

\begin{table*}[t]\renewcommand{\arraystretch}{1.4}
\caption{Comparison of Three Intervention Methods}
\label{tab:intervention-type-description}
\small 
\begin{tabular}{c|ccc}
\hline \hline
\multicolumn{1}{c|}{\multirow{2}{*}{\begin{tabular}[c]{@{}c@{}}\textbf{Intervention} \textbf{Methods}\end{tabular}}} & \multicolumn{3}{c}{\textbf{Characteristics}} \\ \cline{2-4} 
\multicolumn{1}{c|}{} & \multicolumn{1}{c|}{\textbf{Intent Report}} & \multicolumn{1}{c|}{\makecell{\textbf{LLM-powered}\\ \textbf{Persuasion}}} & \makecell{\textbf{Mental-States-Based}\\ \textbf{Persuasion Strategies}} \\ \hline
\textit{Baseline} & \multicolumn{1}{c|}{$\checkmark$} & \multicolumn{1}{c|}{} &  \\
\textit{MindShift-Simple} & \multicolumn{1}{c|}{$\checkmark$} & \multicolumn{1}{c|}{$\checkmark$} &  \\
\textit{MindShift} & \multicolumn{1}{c|}{$\checkmark$} & \multicolumn{1}{c|}{$\checkmark$} & $\checkmark$ \\ 
\hline \bottomrule
\end{tabular}\renewcommand{\arraystretch}{1}
\Description{The table consists of four rows and four columns, including headers. The first column lists three "Intervention Methods": "Baseline," "MindShift-Simple," and "MindShift." The remaining columns describe each method's "Characteristics," specifically "Intent Report," "LLM-powered Persuasion," and "Mental-State-based Persuasion Strategies." The first row is the primary header, and the second row serves as a subheader to specify characteristics. Row three details that "Baseline" has just one characteristic: "Intent Report." Row four shows "MindShift-Simple" has two characteristics: "Intent Report" and "LLM-powered Persuasion." The final row indicates "MindShift" incorporates all three characteristics.}
\end{table*}

\subsection{Experiment Design}
\label{subsub:exp-design}

We adopted a within-subject design, with intervention techniques as independent variables (\textit{Baseline}, \textit{MindShift-Simple}, and \textit{MindShift}).
We designed a 5-week field experiment. To measure users' everyday phone usage behavior, the first week is set as the \textit{Baseline} stage, followed by two weeks of one \textit{MindShift} version and another two weeks of the other version.
We counter-balanced the order of \textit{MindShift-Simple} and \textit{MindShift}.

Our evaluation metrics include various aspects:
(1) Intervention acceptance rate. We measure the percentage of times users accept the intervention and quit the blacklist app use when interventions are shown.
(2) Intervention thumb-up rate. For \textit{MindShift-Simple} and \textit{MindShift} that show persuasion content, users can provide feedback by thumb-up or thumb-down (step 4 in Figure \ref{fig:interaction-process}).
(3) App usage behavior, which includes both app opening frequency and usage duration.
(4) Subjective reports. At the beginning of the study and at the end of each intervention session, we distributed the Smartphone Addiction Scale (SAS)~\cite{kwon2013development} and the self-efficacy scale~\cite{schwarzer1995generalized}. In addition, at the end of the study, we also conducted a brief semi-structured interview to gather user experiences and feedback on our intervention techniques.
These metrics cover both the objective and subjective measures of the interventions.

\subsection{Participants}
\label{subsub:participant}
We sent out recruitment material on social media platforms.
We included a screening survey aiming to identify potential participants who showed signs of smartphone addiction and the willing to reduce their smartphone use.
Specifically, besides basic demographics, we included four questions selected from the SAS and self-efficacy questionnaires, questions about the willingness to reduce smartphone use, the extent of habitual phone use, future plans for the next five weeks, and a screenshot of phone usage time of the last week.
We excluded users (1)  without signs of smartphone addiction (SAS sub-score < 15), or (2) unwilling to reduce smartphone use, or (3) less than 20 hours of weekly phone use, or (4) having special plans such as long-term travel in the next five weeks (which may shift their phone usage patterns).

We received a total of 42 responses. We recruited 31 participants after the screening process. 6 participants voluntarily dropped out during the study.
For the remaining participants, we divided them into groups according to counterbalanced intervention orders.
We conducted a Kruskal-Wallis test analysis to ensure that groups had no significant difference in the SAS scores and self-efficacy scores.
In the end, 25 participants completed the entire study (females=13, males=12, age=22±2 years), including 17 undergraduates, 5 graduate students, and 3 professionals.

\subsection{Experiment Procedure}
\label{sub:exp-procedure}

After all the participants signed the consent form, we held a 20-minute onboarding session online to familiarize participants with the research process and introduce the Android application. We explained in detail the meanings of each selection in the intent report interfaces (Step 2 in Figure \ref{fig:interaction-process}). After the meeting, participants filled out the first SAS and self-efficacy questionnaires. 
The app was then deployed for a 5-week field experiment.

Before users started using one of two versions of \textit{MindShift }(\textit{MindShift} and \textit{MindShift-Simple}), we provided participants with a tutorial explaining the three mental states (\ie boredom, stress, and inertia) that they need to report.
\review{To confirm their understanding, participants were asked to take a brief test on the understanding of mental states until they reached a score of 90\%.}
This ensured that they had an accurate and consistent understanding of the mental states, which in turn improved the accuracy and reliability of the reported data.
At the end of the experiment, participants received a compensation of \$50 for their time.

\section{RESULTS}
During the five-week study, we collected 50,815 minutes of restricted app usage duration, and 54,467 restricted app opening events (7539, 23,769, 21,994, and 835 for habitual use, instrumental use, relax, and quit). 
\review{We conducted statistical tests on the quantitative data collected from the app and scale scores to measure differences. For qualitative data from exit interviews, we conducted thematic coding to extract key insights.}

\subsection{Intervention Acceptance Rate}

The effectiveness of a persuasion strategy is directly measured by the rate of successful prevention of user engagement with the targeted application (\ie intervention acceptance rate). We compared the overall acceptance rate (Sec. \ref{sub:ac-overall}), the acceptance rate for generated persuasion content (Sec. \ref{sub:ac-persua}), and thumb-up rate (Sec. \ref{sub:feedback}). Moreover, we also conducted a detailed analysis of the acceptance rate across strategies (Sec. \ref{sub:ac-stra}), mental states, and activities (Sec. \ref{sub:ac-mental-activity}).

\subsubsection{Overall Acceptance Rate}
\label{sub:ac-round1}

\label{sub:ac-overall}
\begin{figure}[t]
    \centering
     \begin{subfigure}[b]{0.95\linewidth}
         \centering
         \includegraphics[width=\textwidth]{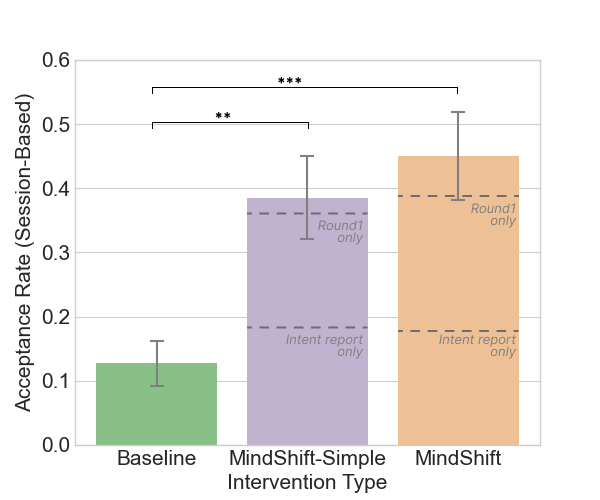}
         \caption{\review{Overall Acceptance Rate (Session-based)}}
         \label{fig:ac-first}
     \end{subfigure}
     \hfill
     \begin{subfigure}[b]{0.95\linewidth}
         \centering
         \includegraphics[width=\textwidth]{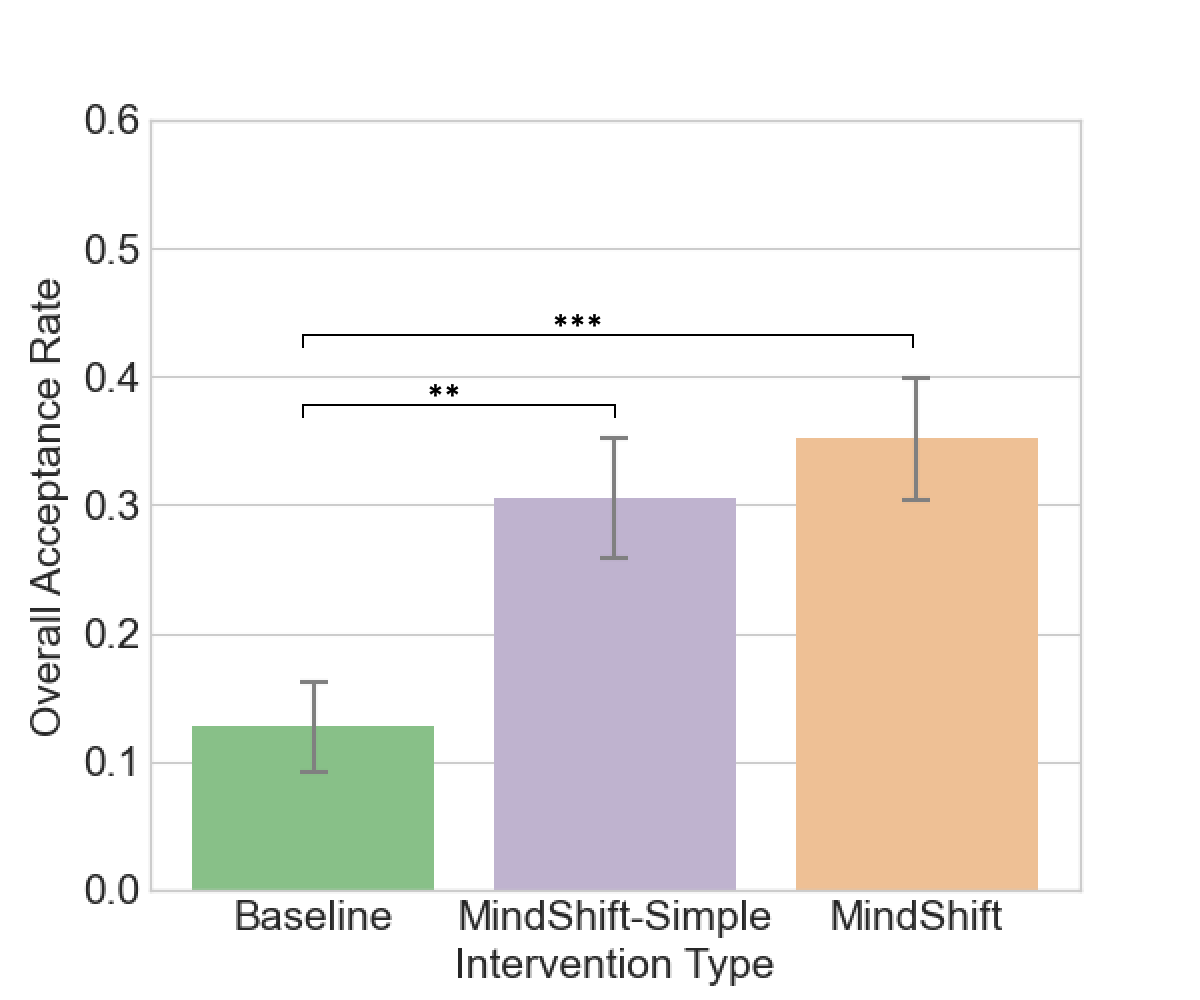}
         \caption{Overall Acceptance Rate (Pop-up-based)}
         \label{fig:ac-all}
     \end{subfigure}
    \caption{Overall Acceptance Rate}
    \label{fig:acceptance}
    \Description{(Top) A bar plot of the session-based intervention acceptance rate of the three techniques. The X-axis is the three techniques. From the left to the right are Baseline, MindShift-Simple, and MindShift. The Y-axis is the acceptance rate(session-based). MindShift has the highest rate at 45.1\%. MindShift-Simple has 38.5\%, and Baseline has 12.7\%. There are also two dashed lines on the bar of MindShift-Simple and MindShift. The upper line represents the first-round acceptance rate. MindShift has the highest rate at 38.7\%, and MindShift-Simple has 36.5\%. The lower line represents the intent-report-only acceptance rate. MindShift-Simple has the highest rate at 18.3\%, and MindShift has 17.9\%.
(Bottom) A bar plot of the pop-up-based intervention acceptance rate of the three techniques. The X-axis is the three techniques. From the left to the right are Baseline, MindShift-Simple, and MindShift. The Y-axis is the acceptance rate(pop-up-based). MindShift has the highest rate at 35.2\%. MindShift-Simple has 30.5\%, and Baseline has 12.7\%.}
\end{figure}

 \textbf{\textit{MindShift} and \textit{MindShift-Simple} increase the overall acceptance rate significantly and \textit{MindShift} achieves best.} 
\review{We assess the overall acceptance rate using two methods. The first ``session-based'' rate means among total app visits (excluding instrumental uses and relaxation), how many times users quit \minorreview{during the intervention (including intent report and persuasion content \footnote{\minorreview{Users click the "I am great and I will quit playing" in Step 1 or "Quit using" in Step 3 in Fig \ref{fig:interaction-process}}}})}. \review{Figure \ref{fig:ac-first} shows that \textit{MindShift} (45.1±34.3\%) achieves higher acceptance than \textit{MindShift-Simple} (38.5±32.2\%) and \textit{Baseline} (12.7±17.4\%).  Significance is observed in a Friedman test ($\chi^2$(2) = 16.64, \textit{p} < .001). Three post hoc Wilcoxon signed-rank tests, corrected with Holm’s sequential Bonferroni procedure, indicate that \textit{MindShift} vs. \textit{Baseline}  (\textit{V} = 31, \textit{p} < .001) and \textit{MindShift-Simple} vs. \textit{Baseline} (\textit{V} = 59, \textit{p} < .01) are significantly different, while \textit{MindShift} vs. \textit{MindShift-Simple} is not (\textit{V} = 126, \textit{n.s.}).}

\review{Considering that the \textit{Baseline} doesn't trigger subsequent interventions like the \textit{Mindshift} and \textit{Mindshift-simple}, we also compare the acceptance rates of the first round in particular. \minorreview{Both \textit{MindShift} and \textit{MindShift-Simple} initiate persuasion immediately after participants report their intents, so we include this initial persuasion in calculating their first-round acceptance rates. Additionally, we analyze the acceptance rate of the intent report to distinguish its effectiveness among the three intervention techniques.}
As shown in the upper dashed lines in Figure \ref{fig:ac-first}, the first-round acceptance rates for \textit{MindShift} (38.7\%±24.9\%) and \textit{MindShift-Simple} (36.2\%±25.8\%) are still significantly higher than the \textit{Baseline} (12.7\%±17.4\%,$\chi^2$(2) = 15.56, \textit{p} < .001).
Post hoc tests show significant differences in \textit{MindShift} vs. \textit{Baseline}  (\textit{V} = 43, \textit{p} < .01), and \textit{MindShift-Simple} vs. \textit{Baseline} (\textit{V} = 32, \textit{p} < .001), but not in \textit{MindShift} vs. \textit{MindShift-Simple} (\textit{V} = 134, \textit{n.s.}).}
\minorreview{
The lower dashed lines in Figure \ref{fig:ac-first} indicate that the acceptance rates when considering only reporting intent are still higher for \textit{MindShift} (17.9\%±17.5\%) and \textit{MindShift-Simple} (18.3\%±18.2\%) compared to the \textit{Baseline} (12.7\%±17.4\%). However, a Friedman test ($\chi^2$(2) = 5.59, \textit{p} < .1) does not show significance, suggesting that intent report has no difference among the three intervention techniques.}

\review{Following the session-based method, we also investigate the pop-up-based acceptance rate, which equals the total number of quit times divided by the total number of intervention pop-ups (i.e., each round is counted as a pop-up).}
Figure \ref{fig:ac-all} shows the comparison, \textit{MindShift} (35.2±24.1\%) still has a higher acceptance than \textit{MindShift-Simple} (30.5±23.6\%) and \textit{Baseline} (12.7±17.4\%, $\chi^2$(2) = 13.69, \textit{p} < .01). Post hoc tests indicate significance for \textit{MindShift} vs. \textit{Baseline}  (\textit{V} = 52, \textit{p} < .01) and \textit{MindShift-Simple} vs. \textit{Baseline} (\textit{V} = 38, \textit{p} < .001), but not for \textit{MindShift} vs. \textit{MindShift-Simple} (\textit{V} = 129, \textit{n.s.}).
\minorreview{Subsequent analyses are all based on pop-ups.}

\subsubsection{Acceptance Rate for Generated Persuasion Content}
\label{sub:ac-persua}

\begin{figure}[]
    \centering
     \begin{subfigure}[b]{\linewidth}
         \centering
         \includegraphics[width=\textwidth]{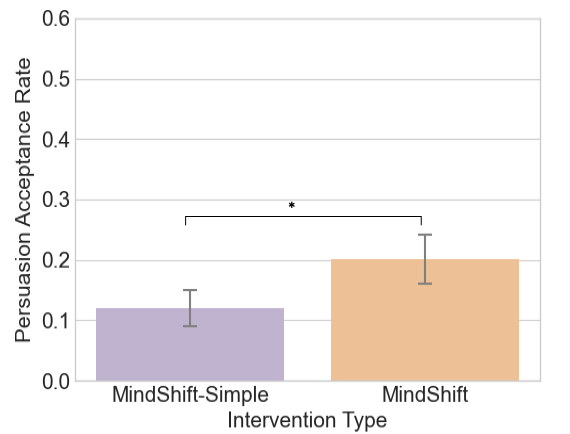}
         \caption{Overall Persuasion Acceptance Rate}
         \label{fig:persuasion-acceptance}
     \end{subfigure}
     \hfill
     \begin{subfigure}[b]{\linewidth}
         \centering
         \includegraphics[width=\textwidth]{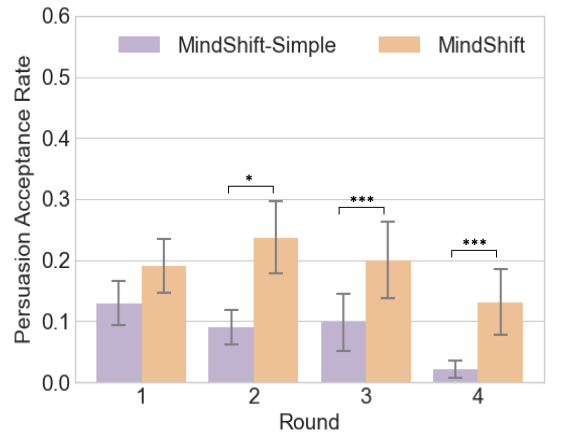}
         \caption{Persuasion Acceptance Rate Grouped by Round}
         \label{fig:ac-round}
     \end{subfigure}
    \caption{Persuasion Acceptance Rate}
    \label{fig:persuasion-acceptance}
    \Description{(Top) A bar plot of the persuasion acceptance rate of two versions of MindShift. The X-axis is the two techniques. From left to right are MindShift-Simple, and MindShift. The Y-axis is the persuasion acceptance rate. MindShift has the highest rate of 20.1\%. MindShift-Simple has 12.0\%.
(Bottom) A bar plot of the persuasion acceptance rate of two versions of MindShift grouped by 4 rounds. The X-axis is the round. From the left to the right are round1, round2, round3, round4. And each round has two bars, the left is MindShift-Simple while the right is MindShift. The Y-axis is the persuasion acceptance rate. MindShift has a higher acceptance rate in every round. }
\end{figure}

\textbf{\textit{MindShift} has a significantly higher persuasion acceptance rate than \textit{MindShift-Simple}.}
The overall acceptance rate includes two parts: exiting when reporting intent and exiting after seeing the persuasion content.
To narrow down the comparison between \textit{MindShift} and \textit{MindShift-Simple}, we \review{exclude the intent report stage and} focus on the acceptance rate during the persuasion stage, as they differ only in the persuasive content.
As shown in Figure \ref{fig:persuasion-acceptance}, \textit{MindShift} achieves higher persuasion acceptance (20.1±20.2\%) than \textit{MindShift-Simple} (12.0±15.0\%) and a paired-samples \textit{t}-test shows that \textit{MindShift} was statistically significantly higher (\textit{t} = -2.21, \textit{p}<0.05).

Moreover, as we introduce in Sec.~\ref{sub:design:when}, every persuasion intervention could consist of 1 to 4 rounds, depending on which mental state participants are in and which stage participants leave the app.
Therefore, we further compare the persuasion acceptance rates between \textit{MindShift} and \textit{MindShift-Simple} across different \minorreview{persuasion} rounds.
The results show that \textit{MindShift} outperforms \textit{MindShift-Simple} at each round \minorreview{\footnote{\minorreview{Round 1 here only contains persuasion and excludes the intent report.}}}($\Delta_{\textit{Round1}}$=6\%, $\Delta_{\textit{Round2}}$=14.7\%, $\Delta_{\textit{Round3}}$=10.2\%, $\Delta_{\textit{Round4}}$=11\%) as indicated in  Figure \ref{fig:ac-round}.
We conduct a paired-samples \textit{t}-test between the two intervention techniques in each round. Results indicate that, except for the first round, rounds 2 (\textit{p}<0.05), 3  (\textit{p}<0.001), and 4 (\textit{p}<0.001) all exhibit that \textit{MindShift} achieves significantly higher acceptance rate. \review{This trend suggests that as the number of interventions increases, \textit{MindShift}'s advantage becomes more pronounced, highlighting \textit{MindShift}'s robustness and effectiveness in maintaining high acceptance rates.}

\subsubsection{Thumb-up Rate of Interventions}
\label{sub:feedback}
\begin{figure}
\centering
    \begin{subfigure}{\linewidth}
    \centering
    \includegraphics[width=1\linewidth]{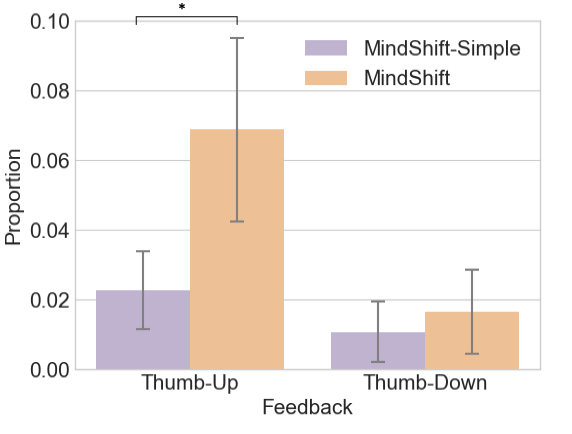}
    \caption{Feedback Proportion}
    \label{fig:feedback}
    \end{subfigure}
    \begin{subfigure}{\linewidth}
    \centering
    \includegraphics[width=1\linewidth]{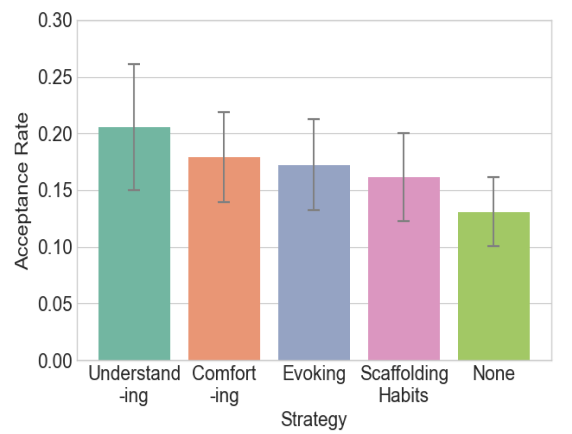}
    \caption{Acceptance Rate Grouped by Persuasion Strategies}
    \label{fig:ac-strategy}
    \end{subfigure}
    \caption{Feedback and Strategy Acceptance Rate}
    \Description{(Top) A bar plot of the feedback proportion of two versions of MindShift. The X-axis is the two feedback categories. From the left to the right are Thumb-Up, and Thumb-Down, and each category has two bars,  the left is MindShift-Simple while the right is MindShift. The Y-axis is the proportion rate. The thumb-up proportion for MindShift-Simple is 2.2\% while MindShift is 6.8\%. The thumb-down proportion for MindShift-Simple is 1.6\% while MindShift is 2.3\%. (Bottom) A bar plot of the persuasion acceptance rate of different strategies. The Y-axis is the acceptance rate.  The X-axis is the strategy. From the left to the right are understanding, comforting, evoking, scaffolding habits, and MindShift-Simple, they have acceptance rates of 20.9\%, 15.3\%, 22.3\%, 16.9\%, and 12.0\% respectively. Each strategy has a higher acceptance rate than MindShift-Simple.}
\end{figure}
\textbf{\textit{MindShift} has a significantly higher thumb-up rate.} Users can give feedback in the persuasion interface. As depicted in Figure \ref{fig:feedback},  6.8\% of interventions in \textit{MindShift} receives thumb-up while \textit{MindShift-Simple} receives only 2.2\%.
A paired-samples \textit{t}-test shows that significant differences (\textit{p}<0.05) are observed for the thumb-up rate, but no significant differences (\textit{p}=0.22) for the thumb-down rate between the two techniques. This indicates that \textit{MindShift} aligns better with users' preferences.

\subsubsection{Acceptance Rate across Different Strategies}
\label{sub:ac-stra}
We design four persuasion strategies in \textit{MindShift} whereas \textit{MindShift-Simple} does not incorporate any specific strategies, so we further compare the persuasion acceptance rates across different strategies. Figure \ref{fig:ac-strategy} shows that all strategies we design outperform \textit{MindShift-Simple}  
($\Delta_{\textit{Understanding}}$=8.9\%, $\Delta_{\textit{Comforting}}$=3.3\%, $\Delta_{\textit{Evoking}}$=10.3\%,
$\Delta$~\allowbreak$_{\textit{Scaffolding}}$~\allowbreak$_{\textit{Habits}}$=4.9\%)
but the differences are not statistically significant.

\begin{figure}
    \centering
     \begin{subfigure}[b]{\linewidth}
         \centering
         \includegraphics[width=\textwidth]{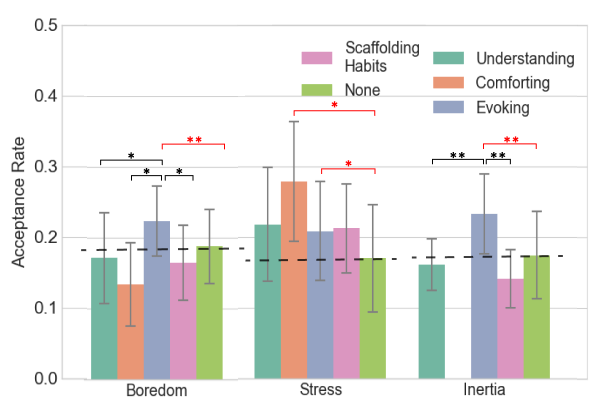}
         \caption{Strategy Acceptance Rate Grouped by Mental State. Significant differences compared to no strategy (green bar) are highlighted in red, while differences among strategies are indicated in black.}
         \label{fig:ac-emotion}
     \end{subfigure}
     \hfill
     \begin{subfigure}[b]{\linewidth}
         \centering
         \includegraphics[width=\textwidth]{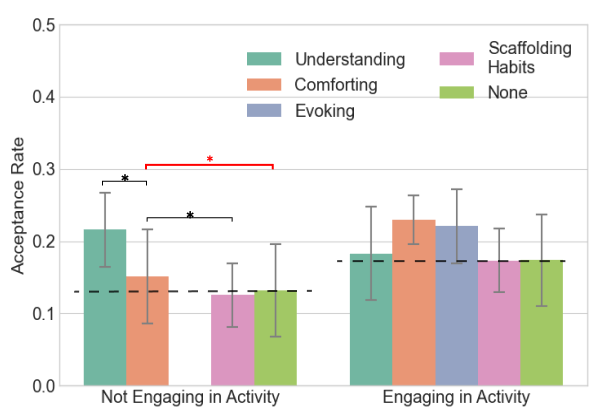}
         \caption{Strategy Acceptance Rate Grouped by Activity. The same annotation method as grouped by mental state.}
         \label{fig:ac-activity}
     \end{subfigure}
    \caption{Strategy Acceptance Rate Grouped by Different Mental States}
    \Description{(Top) A bar plot of the strategy acceptance rate grouped by 3 mental states. The Y-axis is the acceptance rate. The X-axis is the mental state. From the left to the right are stress, inertia, and boredom. Each mental state has five bars, from left to right are understanding, comforting, evoking, scaffolding habits, and MindShift-Simple. Compared with MindShift-Simple, under the mental state of "Stress", every strategy has a higher acceptance rate. Under the mental state of Inertia" and "Boredom", Evoking is higher.
(Bottom) A bar plot of the strategy acceptance rate grouped by 2 activity states. The Y-axis is the acceptance rate. The X-axis is the activity state. From the left to the right are not engaging in activity and engaging in activity. Each activity state has five bars, from left to right are understanding, comforting, evoking, scaffolding habits, and MindShift-Simple. Compared with MindShift-Simple, when not engaging in activity, understanding and comforting have a higher acceptance rate. when engaging in activity, all strategies have a higher acceptance rate.}
\end{figure}

\subsubsection{Strategy Acceptance Rate across Different Mental States and Activities}
\label{sub:ac-mental-activity}
As we show in Figure~\ref{fig:strategy}, each mental state has a different strategy mapping. Therefore, we also seek to derive insights regarding which strategies are most effective for users under different mental states (Figure \ref{fig:ac-emotion}) and activities (Figure \ref{fig:ac-activity}). 
Friedman test and post hoc Wilcoxon signed-rank tests are employed to investigate the influence of different strategies on acceptance rate.

For mental states: 
    under the mental state of \textit{"Boredom"} and \textit{"Inertia"}, \textit{Evoking} is significantly more effective ($\Delta_{\textit{Boredom}}$=3.6\%, $\Delta_{\textit{Inertia}}$=5.8\% ) than \textit{MindShift-Simple} (\textit{ps} < 0.01);
    under the mental state of \textit{"Stress"},  \textit{Comforting} and \textit{Evoking} show a trend toward significance compared with \textit{MindShift-Simple} ($\Delta_{\textit{}}$ = 10.8\%,  \textit{p}<0.1 and $\Delta_{\textit{}}$ = 3.8\%,  \textit{p}<0.1 respectively).
    
For activity levels:
    when not engaging in activity,  \textit{Comforting} is significantly more effective than \textit{MindShift-Simple}  ($\Delta_{\textit{}}$=5.6\%, \textit{p} < 0.05);
    when engaging in activity, there are no significant differences in all the strategies compared to \textit{MindShift-Simple}.

\subsection{App Usage Behavior}
We then investigate the influence of the intervention on participants’ app usage behavior. Overall, participants have less app usage frequency and duration when using \textit{MindShift} and \textit{MindShift-Simple}, especially in habitual usage.

\subsubsection{Overall Usage Behavior}
\label{subsub:res-usage}
\begin{figure}[t]
    \centering
     \begin{subfigure}[b]{0.9\linewidth}
         \centering
         \includegraphics[width=\textwidth]{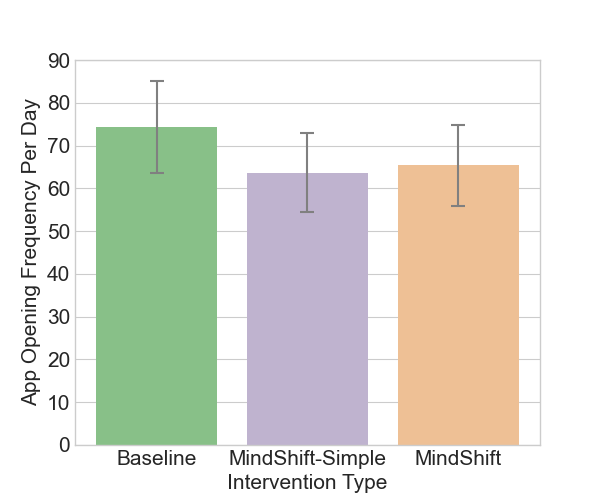}
         \caption{Total App Opening Frequency}
         \label{fig:total-frequency}
     \end{subfigure}
     \hfill
     \begin{subfigure}[b]{0.9\linewidth}
         \centering
         \includegraphics[width=\textwidth]{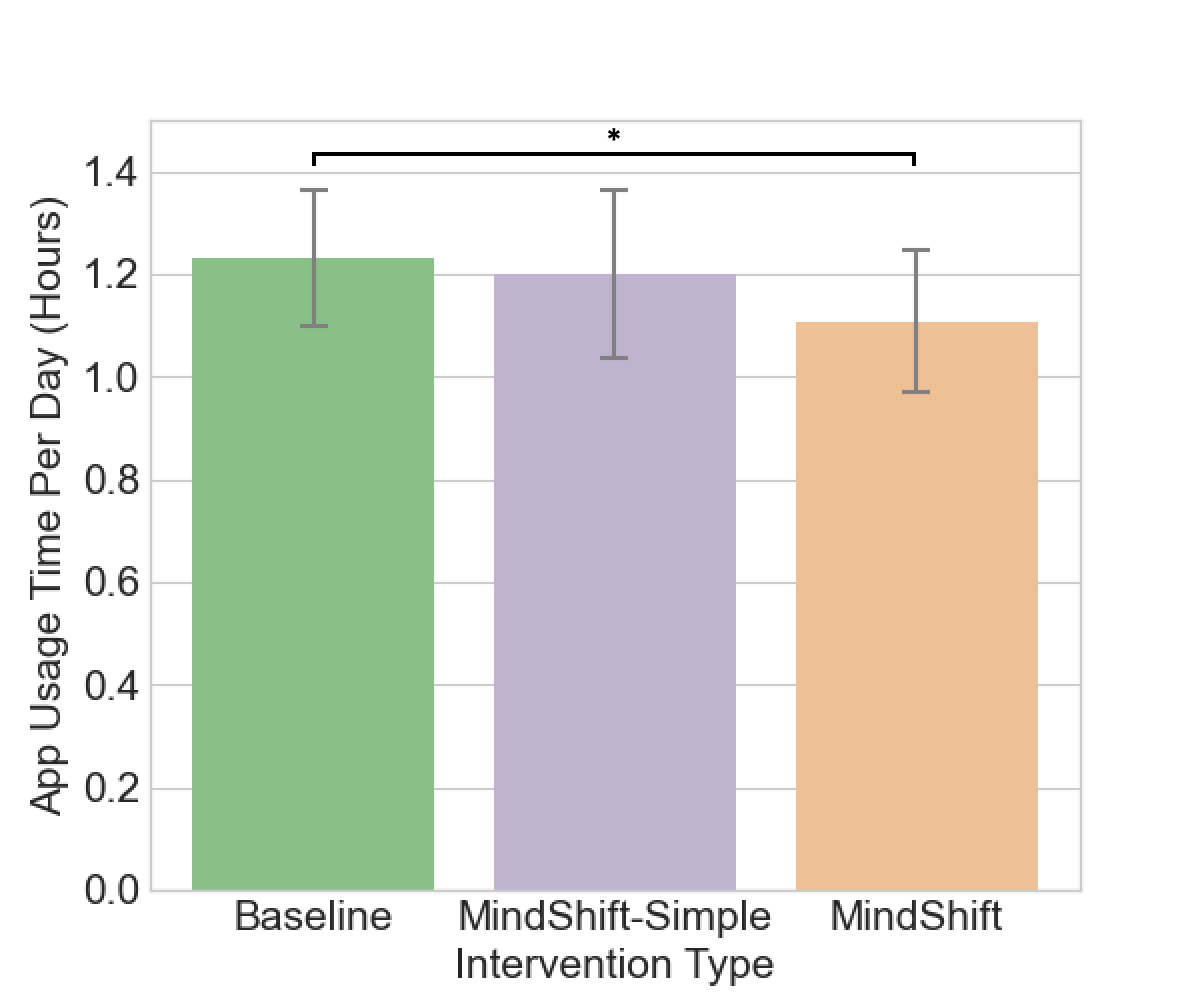}
         \caption{Total Phone Usage (Hours) per Day}
         \label{fig:total-usage}
     \end{subfigure}
    \caption{App Usage Behavior}
    \Description{(Top) A bar plot of the app opening frequency of all restricted apps of the three techniques. The X-axis is the three techniques. From the left to the right are Baseline, MindShift-Simple, and MindShift. The Y-axis is the opening frequency per day. MindShift-Simple (63±9.2) and MindShift (65±9.5) had lower opening frequency than Baseline (74±9.7).
(Bottom) A bar plot of the app usage duration of all restricted apps of the three techniques. The X-axis is the three techniques. From the left to the right are Baseline, MindShift-Simple, and MindShift. The Y-axis is the usage duration per day. MindShift (1.11 ±0.7 hours) has the lowest app usage duration compared to the Baseline (1.23±0.7 hours) and MindShift-Simple(1.20 ± 0.8 hours).}
\end{figure}

We count the number of app opening attempts for restricted apps. Figure \ref{fig:total-frequency} presents the opening frequency (daily open count) under three intervention techniques. \textit{MindShift-Simple} (63.6±9.2) and \textit{MindShift} (65.3±9.5) have lower opening frequency than \textit{Baseline} (74.3±9.7). Compared to \textit{Baseline}, \textit{MindShift} reduces by 12.1\% usage duration while \textit{MindShift-Simple} reduces by 14.4\%. However, a Friedman test does not show significance.

We also measure restricted app usage duration, another important factor for phone overuse.  As can be seen from Figure \ref{fig:total-usage}, participants have the lowest app usage duration in \textit{MindShift} (1.11 ±0.7 hours) compared to the \textit{Baseline} (1.23±0.7 hours) and \textit{MindShift-Simple} (1.20 ± 0.8 hours). Compared to \textit{Baseline}, \textit{MindShift} reduces by 9.8\% usage duration while \textit{MindShift-Simple} reduces by 2.4\%. We conduct a Friedman test and find a significant difference among different techniques (\textit{p}<0.05). A post-hoc Wilcoxon test shows that \textit{MindShift} has a trend of declining compared to \textit{Baseline}, with marginal significance (\textit{p}<0.1). 

\review{To validate the assumption that the intent and mental state remain stable in one unlock session across different apps in Sec.~\ref{sub:design:when}, we analyze the duration of users' intent and mental states.
The results show that the median duration of a mental state is 5 hours (third quartile 14.5 hours). The median duration for an intent (changing from habitual use to other intents) is 37 minutes (third quartile 60 minutes). This validates our hypothesis that intent and mental state are stable during one habitual usage session. 
}

\subsubsection{Habitual Usage Behavior}
\label{subsub:habitual}

\textbf{\textit{MindShift} and \textit{MindShift-Simple} significantly reduce habitual app usage duration and frequency.}
The focus of our intervention is habitual use, so we investigate the changes in habitual usage behavior.  Results show that \textit{MindShift} and \textit{MindShift-Simple} can both significantly reduce habitual use. 
The app visit frequency and duration of habitual usage cases also decrease significantly during the two versions of \textit{MindShift} compared to \textit{Baseline} ($\Delta_{\textit{\textit{MindShift-Simple}}}$ equaled 80.6\% for visit frequency, 84.4\% for usage duration, and 6.8\% for habitual use proportion, $ps$ < 0.001; $\Delta_{\textit{MindShift}}$ equaled 77.3\% for visit frequency, 80\% for usage duration and 6.8\% for habitual use proportion, $p$s < 0.001).

\review{To confirm the suitability of the 2-minute intervention interval, we analyze data on users' habitual usage duration during the \textit{Baseline} phase (unaffected by subsequent interventions).  Analysis shows that 75\% of users spent about 4 minutes in a single habitual use, supporting our design choice of the 2-min intervention interval.}

\subsection{Subjective Report}
We further analyze on user-reported SAS and self-efficacy scale results. When using \textit{MindShift }and\textit{ MindShift-Simple}, participants experience a significant decrease in SAS score and a significant increase in self-efficacy score, but they see no change when using \textit{Baseline}. We summarize the results as follows.

\begin{figure}[t]
    \centering
     \begin{subfigure}[b]{0.95\linewidth}
         \centering
         \includegraphics[width=\textwidth]{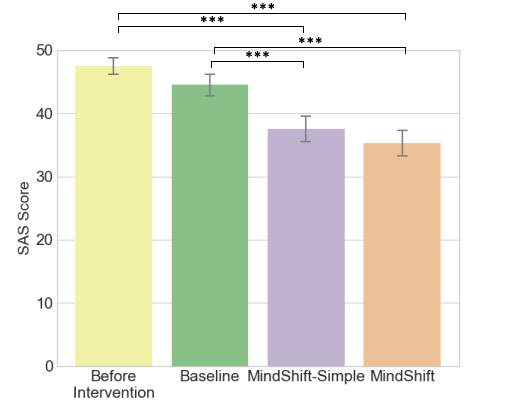}
         \caption{SAS Score}
         \label{fig:sas}
     \end{subfigure}
     \hfill
     \begin{subfigure}[b]{0.95\linewidth}
         \centering
         \includegraphics[width=\textwidth]{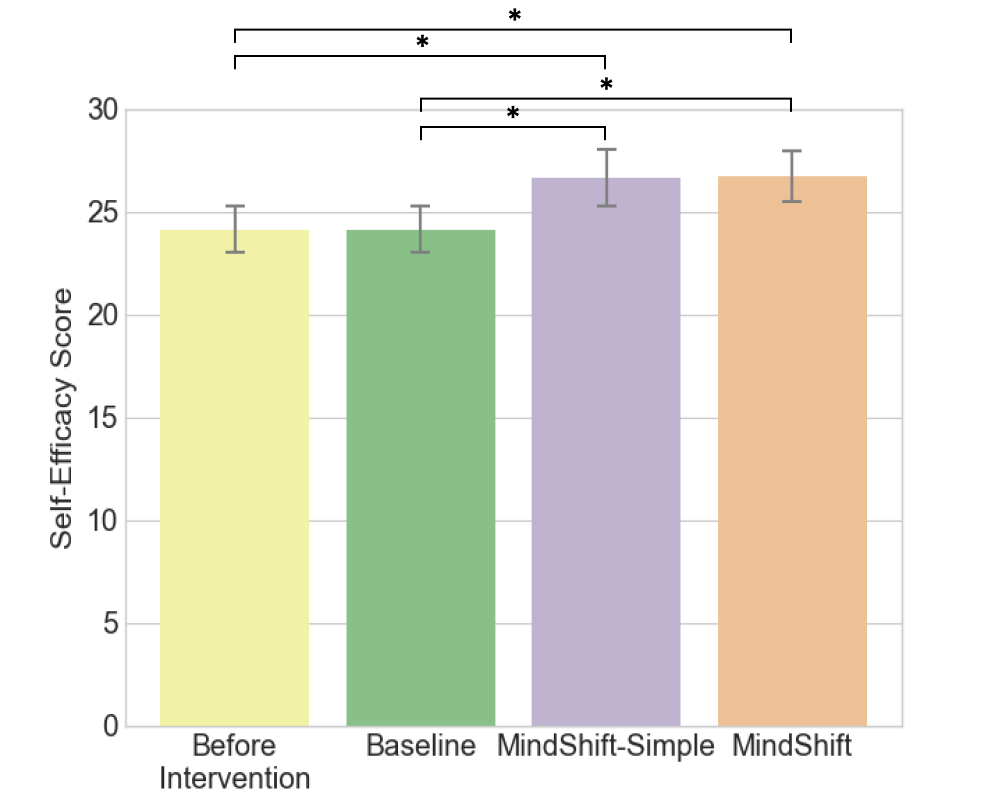}
         \caption{Self-efficacy Score}
         \label{fig:efficacy}
     \end{subfigure}
    \caption{Subjective Scales Report}
    \Description{(Top) A bar plot of the SAS score of the four intervention stages. The X-axis is the four stages. From the left to the right are Before Intervention, Baseline, MindShift-Simple, and MindShift. The Y-axis is the SAS Score.  MindShift exhibited the lowest SAS scores (35.2±10.2), followed by MindShift-Simple in the second position (37.6±10.1), with the Baseline intervention ranking the third (44.5±8.7) and the initial ranking the last (47.5±6.7).
(Bottom) A bar plot of the Self-efficacy score of the four intervention stages. The X-axis is the four stages. From the left to the right are Before Intervention, Baseline, MindShift-Simple, and MindShift. The Y-axis is the Self-efficacy Score.  MindShift(26.7±6.1) and MindShift-Simple(26.6±6.8) exhibited higher scores compared to the Baseline intervention (24.1±5.6)  and the initial (24.1±5.6). }
\end{figure}

\subsubsection{Decrease of SAS Score}
\textbf{\textit{MindShift} and \textit{MindShift-Simple} decrease SAS score significantly and \textit{MindShift} achieves best}.
Figure \ref{fig:sas} shows the results of the SAS scores during
the intervention stages. \textit{MindShift} exhibits the lowest SAS scores (35.2±10.2), followed by \textit{MindShift-Simple} in the second position (37.6±10.1), with the \textit{Baseline} intervention ranking the third (44.5±8.7) and the initial ranking the last (47.5±6.7). This indicates that\textit{ MindShift} and \textit{MindShift-Simple} reduce SAS scores by 34.7 and 25.8\%. The results of a Friedman test show a significant difference (\textit{p}<0.001). Post-hoc Wilcoxon tests show that both \textit{MindShift} and \textit{MindShift-Simple} are significantly lower than the \textit{Baseline} and initial scores (all \textit{p}s<0.01). This suggests that the two versions of \textit{MindShift} have the potential to fundamentally alter individuals' mobile phone usage behavior.

\subsubsection{Increase of Self-efficacy Score}
\textbf{\textit{MindShift} and \textit{MindShift-Simple} increase Self-efficacy score significantly}.
Figure \ref{fig:efficacy} shows the results of the self-efficacy scores during the intervention stages. \textit{MindShift} (26.7±6.1) and \textit{MindShift-Simple} (26.6±6.8) exhibit higher scores compared to the \textit{Baseline} intervention (24.1±5.6)  and the initial (24.1±5.6). This indicates that \textit{MindShift} and \textit{MindShift-Simple} increase self-efficacy scores by 10.7\% and 10.4\%. Friedman test shows a significant difference (\textit{p}<0.001). The post-hoc Wilcoxon test shows that both \textit{MindShift} and \textit{MindShift-Simple} are significantly higher than the \textit{Baseline} and initial scores (all \textit{p}s<0.05). This indicates that persuasive techniques have the potential to enhance individuals' self-efficacy, which can result in overcoming excessive mobile phone usage. In contrast, conventional reminders may not achieve this goal.

\subsubsection{Subjective Comments}

During the exit interview, participants generally had a positive experience.
One participant said, \textit{``I can feel that my recent dependency on the phone has decreased''} (P17). One participant felt using the app shifted them to self-improvement tasks instead of mindlessly usage, ``\textit{Now, when I have nothing to do, I tend to do other things, like learning vocabulary, instead of aimlessly browsing my phone''} (P10). \review{Another participant was willing to use it longer, \textit{``I'm a little sad with the disappearance of pop-ups after the experiment. If possible, I would like to keep using it"} (P3).}
Additionally, participants had positive comments on \textit{MindShift} which includes the mental states factor, ``\textit{I feel that its suggestions align well with my emotional state at that time.''} (P25).
They also valued the ``\textit{Understanding}'' and ``\textit{Comforting}'' strategies, saying ``\textit{It tells me that I'm not the only one experiencing these painful emotions, which helps me feel better}'' (P22).

Despite the majority of positive comments, a small number of participants expressed their dissatisfaction with \textit{MindShift}.
 Some participants found the persuasive message to be ``\textit{a bit stiff and templated}'' (P4, P6), and they believed that ``\textit{they would develop tolerance as they repeatedly use''} (P20). Some participants also mentioned privacy concerns. P15 mentioned that \textit{MindShift-Simple}'s ability to capture time and location made her uncomfortable.
\review{Moreover, participants’ preference for linguistic characteristics is highly personal. Some felt harsh ones were more useful, \textit{``Gentle tone doesn't work for me, I wish the words could be harsher"} (P8). Some preferred data proof than pure textual reasoning, \textit{``It's intuitive to tell me how long I have used my phone today directly. The number is very eye-catching"} (P9). This suggests the future direction of personalized persuasive content design. We have more discussion in Sec.~\ref{sub:discussion:adaptive}.}

\subsection{Summary of Results}
Overall, two versions of \textit{MindShift} show significantly higher acceptance rates compared to the \textit{Baseline}. \textit{MindShift} has the highest acceptance \review{(45.1\% for session-based and 35.2\% for pop-up-based)} and thumb-up rates (6.8\%), and it statistically significantly outperforms \textit{MindShift-Simple} in both acceptance rates of generated persuasion content (8.1\%) and on a per-round basis (6-14.7\%). Furthermore, there exist strategies that significantly outperform \textit{MindShift-Simple} (3.6-10.8\%) in every mental state, suggesting that the strategies we design are meaningful. 
Furthermore, \textit{MindShift }and \textit{MindShift-Simple} lead to a decrease in overall app opening frequency (12.1-14.4\%) and usage duration (9.8-2.4\%) and are significantly effective in reducing habitual use. \textit{MindShift} and \textit{MindShift-Simple} also reduce SAS scores by 34.7-25.8\% and increase self-efficacy scores by 10.7-10.4\% statistically significantly while \textit{Baseline} does not. This suggests that \textit{MindShift} has the potential to profoundly transform human behavior with enduring effects.
Finally, users' subjective comments also confirm a perceived reduction in smartphone dependency and an inclination to continue using \textit{MindShift}.

\section{DISCUSSION}
\label{sec:discussion}

In this section, we discuss \textit{MindShift}'s novelty in contrast to previous intervention techniques (Sec \ref{sub:discussion:roles}), future work (Sec \ref{sub:discussion:adaptive}), the potential of leveraging LLMs for behavior change (Sec \ref{sub:discussion:llms}), and the limitations (Sec \ref{sub:discussion:limitation}).

\subsection{The Roles of Users' Phone Use Purpose and Mental States in Smartphone Intervention}
\label{sub:discussion:roles}
Most previous intervention techniques initiate interventions based on the amount of time and frequency of smartphone usage. However, quantifying smartphone use just by time oversimplifies and ignores the underlying causes. People use smartphones for work, study, and relaxation, as long as for meaningful reasons, the time is not a true problem. \review{As Lukoff et al. found, even if participants didn't reduce their screen time, the intervention could make them feel better in the sense of agency and goal alignment, indicating users prioritize the quality of time over quantification ~\cite{lukoff2023switchTube}.} \textit{MindShift} initiates persuasion only when users recognize their current usage as habitual, aiming to enhance users' self-awareness of their habitual phone use behavior.

Additionally, certain mental states are linked to habitual smartphone use as a form of self-distraction. Although smartphone use serves as a coping mechanism for emotion fluctuations, studies show that habitually using them for escapism fails to effectively mitigate emotions~\cite{duvenage2020technology}. Using smartphones for emotional regulation can lead to problematic smartphone use behavior, potentially leading to severe psychological issues such as depression~\cite{coyne2021tantrums,zsido2021role}. \textit{MindShift} aims to intervene in habitual smartphone usage triggered by specific mental states. Our goal is to reduce users' problematic smartphone use and help individuals transition from avoidance-oriented coping to approach-oriented coping~\cite{compas2001coping}.

\subsection{Towards Adaptive Persuasion Intervention}
\label{sub:discussion:adaptive}
\textit{MindShift} generates dynamic and personalized persuasion content by combining information such as users' simple physical contexts, mental states, and other behaviors. However, several participants still mentioned that the LLM-generated content sometimes could be ``\textit{stiff and templated}''.
This may be attributed to the limited prompt templates. Although the content generated by the LLM varies, the main theme is guided by our prompts, which could limit the variation of persuasion content.
To improve, we suggest integrating user feedback into the system for more adaptive intervention. Currently, \textit{MindShift} supports a simple thumb-up and thumb-down feedback mechanism. Even with such simple information, we could establish a human-in-the-loop setup to fine-tune content, aligning better with user preferences.
Another aspect is to include more diverse behavior features captured by passive sensors on smartphones and wearables~\cite{xu_globem_2022,meegahapola2023generalization,he2020pneufetch}.

Moreover, future work can also consider collecting more comprehensive feedback from users. Users could customize the language style generated by an LLM, which can be coupled with adaptive algorithms such as reinforcement learning to achieve a more intelligent just-in-time adaptive intervention (JITAI) system that evolves with users ~\cite{goldstein2017return,orzikulova2024time2stop}.

\subsection{Leveraging LLMs for Behavior Change}
\label{sub:discussion:llms}
\subsubsection{Advantages of Using LLMs for Behavior Change}

\review{Our study sheds light on the possibility of leveraging LLMs to change user behavior by influencing human cognition.} 
\review{Previous efforts in this field often hinge on users' ability to self-reflect and self-persuade, limited by the narrow scope of sentence databases used~\cite{xu2022typeout}. LLMs break this barrier, offering a broader range of persuasive strategies. They can generate adaptive and diverse persuasive content, tailored to the individual's context. In our study, we observed notable changes in cognition: participants' smartphone addiction scale scores dropped, and self-efficacy scores rose after the study (Figure \ref{fig:sas}). This suggests a potential for long-term behavioral change, which we aim to explore further in our future work. }

\review{We also want to highlight that \textit{MindShift} is just one example of the possibilities in this domain. Future research could integrate LLMs for more dynamic interventions, such as self-affirmation content generation in the typing intervention~\cite{xu2022typeout} and personalized visualizations~\cite{hiniker2016mytime}. Moreover, our methodology can potentially be expanded to other domains, such as smoking or alcohol cessation, eating diet, and physical activity promotion, where LLMs can be used to generate context-aware dynamic persuasive content for a specific well-being goal. We envision our work can inspire a number of creative LLM-powered intervention techniques in the future.}
\review{
}

\subsubsection{Design Implication for Using LLMs in Other Behavior Change Domains}
\label{subsub:implication}
\review{
Based on our findings in the study, we extract three design implications of using LLMs for behavior change in various domains. 
} 

\review{First, investigating why people behave in certain ways is the basis of any intervention design, especially when LLMs can utilize such insights when generating persuasion. Our study explores the psychological factors behind habitual smartphone use. \textit{MindShift} leveraging those factors outperformed \textit{MindShift-Simple} only considering physical factors (see in Sec. \ref{sub:ac-round1} to \ref{sub:feedback}).}

\review{Second, context is crucial for LLMs to generate dynamic, tailored persuasive messages.
Our examples in Table~\ref{tab:prompt-comparsion} showcase the importance of user contexts for content generation.
In our study, some user contexts, like mental states, cannot be detected automatically but depend on users' self-reports, which can be improved in future design. 
Past work has explored how to use smartphone usage data and machine learning to predict boredom and stress~\cite{pielot2015attention,lekkas2022using,ciman2016individuals,stutz2015smartphone,meegahapola2023generalization}, and there have been researches using physiological measuring instruments to learn mental states from biosignals~\cite{seo2019machine,seo2019exploration,xu2019leveraging}. With the development of more smart and wearable technologies, there is the potential to track users' mental states automatically~\cite{hickey2021smart,gedam2021review}. This can simplify the intervention process and improve user experience. However, it remains an open question on how skipping self-reflection may impact the effectiveness of such a persuasion technique.}

\review{Last, crafting a suitable prompt is crucial for effectively incorporating expert knowledge into LLM generation. This often involves multiple attempts and adjustments to ensure the generated content aligns with expectations.
While not the main contribution of our work, we conducted extensive iterations to ensure the appropriateness of the persuasive content. We have more discussion on the ethical concerns if prompt engineering is not done properly in the next paragraph.
We refer future developers to recent studies, such as \textit{EmotionPrompt}~\cite{li2023emotionprompt}, for more comprehensive guidance on enhancing LLM outputs.
}

\subsubsection{Ethical Concerns and Risk of Using LLMs for Behavior Change}
\label{subsub:LLMRisk}
Although \textit{MindShift} performs well in changing problematic smartphone use, there are important ethical concerns we want to highlight about the risk of using LLMs for behavior change. 

Despite carefully crafted prompts, developers face challenges ensuring the constant safety of generated persuasive messages. \review{For example, while we fixed hallucination for our experiment, it is one of the biggest concerns in LLMs and can still possibly occur in real-world deployment~\cite{ji2023survey,ferrer2021bias}.
Additionally, although we didn't encounter it in our experiment, LLMs can generate dangerous content, such as abusive and discriminatory sentences.}
Furthermore, there is still room to improve LLMs' understanding of the nuances of human mental states~\cite{xu2023mentalllm}. For instance, if users of \textit{MindShift} are already stressed due to their life objectives (\eg struggling with academic stress), additional reminders of these goals could exacerbate the stress or even cause harm.
\review{
More future work is needed to improve safety and reduce the risk before we deploy LLMs for large-scale intervention studies.}

Moreover, privacy is another critical concern since the detection of users' physical and psychological context data is needed. \textit{MindShift} employs a commercial API from OpenAI, transmitting users' data to a third party. Although we intentionally designed the prompt to avoid including any identifiable information, there is still the risk of revealing information about their behavior and mental states.
One solution for future study is to leverage open-sourced LLMs (such as LLaMA2~\cite{touvron2023llama} or PaLM2~\cite{anil2023palm}) so that user's data can be appropriately handled and encrypted by ourselves instead of a third party.

\subsection{Limitations}
\label{sub:discussion:limitation}
Our study has a few limitations.
First, our experimental user group is limited to young adults, limiting result applicability. Future studies can expand the sample size and involve more diverse user groups.
Second, our field experiment is short. A five-week deployment cannot reveal the longitudinal effect of such an intervention technique. Moreover, if our time and monetary budget allow, our experiment design can be improved by making the \textit{Baseline} another two-week intervention session for a more fair comparison.
\review{Third, the validity and reliability test of mental state report questions needs further improvement. Testing convergent and discriminant validity, recruiting more samples for reliability tests, and using statistical methods to evaluate consistency are areas for enhancement.}
Additionally, our current method for detecting habitual use and mental states relies on self-reporting, increasing users' burden. We only consider initial habitual use upon users unlocking phones, neglecting shifts in user purposes during app usage. As we mentioned in Sec.~\ref{subsub:implication}, future work can explore automatic intent and mental state detection, and the data collected in this study can serve as a starting point for machine learning training models.
Finally, our study utilizes GPT-3.5 as a large language model, and its performance is still unstable.
Future work can explore more lighted weighted and robust LLMs for local deployment, which can address the concerns mentioned in Sec.~\ref{subsub:LLMRisk} to some extent.
\section{CONCLUSION}

This paper introduces \textit{MindShift}, a mental-based persuasion intervention technique powered by LLM designed to mitigate problematic smartphone use. We conducted a Wizard-of-Oz study and an interview study to explore the mental states behind problematic smartphone use: \textit{stress}, \textit{boredom}, \textit{inertia}, and designed four persuasion strategies: \textit{understanding}, \textit{comforting}, \textit{evoking}, and \textit{scaffolding habits}. \textit{MindShift} (1) collects users' usage intent, usage behavior, physical context, mental states, goals\&habits, (2) uses the persuasion strategies we design, (3) leverages LLMs to generate dynamic, personalized persuasion messages. 
Through a five-week within-subjects user experiment (N=25), we compared three intervention techniques (\textit{MindShift}, \textit{MindShift-Simple}, \textit{Baseline)}. \textit{MindShift} outperforms \textit{MindShift-Simple} and \textit{Baseline}, improving acceptance rates (4.7-22.5\%) and reducing app usage (7.4-9.8\%). 
Notably, \textit{MindShift} and \textit{MindShift-Simple} significantly reduce SAS scores (34.7-25.8\%) and increase self-efficacy scores (10.7-10.4\%). Finally, users' subjective comments also confirm a perceived reduction in smartphone dependency and a willingness to continue to use \textit{MindShift}.
Our work provides valuable insights into the mental states behind problematic smartphone use and the effectiveness of LLMs-powered persuasion for smartphone intervention.


\bibliographystyle{ACM-Reference-Format}
\bibliography{sample-base}

\appendix\onecolumn
\section*{Appendix}

\begin{table*}[h]
\centering
\small
\caption{\review{Persuasive Messages in WoZ Study. Four types of persuasive messages delivered in WoZ study and their examples.}}
\label{tab:woz-message}
\resizebox{0.75\textwidth}{!}{%
\begin{tabular}{ll}
\toprule
\multicolumn{1}{c}{\textbf{Types}} & \multicolumn{1}{c}{\textbf{Examples}} \\
\midrule
Usage Notice       & \textit{"You have used Wechat for 2 hours and 25 minutes today. Put down your phone please!"} \\
Practical Guidance & \textit{"Are you still using WeChat? Have you completed the task of analyzing data today?"} \\
Encouragement      & \textit{"You have only spent 2 hours on your phone today, that's excellent! Keep up the good work}$\sim$\textit{"} \\
Deterrent          & \textit{"Using the phone before bedtime can affect the quality of your sleep."} \\
\bottomrule
\end{tabular}%
}
\end{table*}

\begin{table*}[h]
\centering
\caption{\review{Takeaways from WoZ \& semi-structured interview studies. (E) shows messages sent by experimenters, (W) demonstrates participants' reflection quotes in the WoZ study, and (S) means participants' quotes in the semi-structured interview study.}}
\label{tab:takeaways}
\small 
\begin{tabular}{p{0.22\textwidth} p{0.74\textwidth}}
\toprule
\multicolumn{2}{c}{\textbf{WoZ study}} \\
\midrule
\textbf{Types of smartphone use} & \textbf{Representative quote(s)} \\
\cline{1-2}
Instrumental use\newline (not to be intervened) & \textit{"You've already spent one and a half an hour on WeChat today. Please put your phone down and focus on other aspects of life. (E)"}\newline
  \textit{"I felt a bit resentful because I was using WeChat to manage my affairs, rather than idly wasting time. (W1)"}\\
\addlinespace
Instrumental use - relaxation\newline (not to be intervened) &
  \textit{"Please stop browsing Zhihu and engage in more meaningful activities. (E)"}\newline
  \textit{"I don't agree. Finding joy in browsing Zhihu constitutes meaning for me. (W5)"}\\
\addlinespace
Habitual use \newline(to be intervened) &
  \textit{"You have used Zhihu for 2 hours today. Think about what else you have to do tonight. (E)"}\newline
  \textit{"Thanks for this suggestion. I always failed to control myself to open Zhihu. (W10)"}\\
\hline
\textbf{Factors affecting persuasion effectiveness} & \textbf{Representative quote(s)} \\
\hline
Mental states &
  \textit{"When you find yourself with idle time, consider engaging in meaningful activities such as reading, writing, or drawing. (E)"}\newline
  \textit{"It correctly identified my state of not knowing what to do and offered sensible advice. (W11)"}\\
\addlinespace
Personal goals &
  \textit{"Your WeChat session has lasted 10 minutes. Please set aside your device to alleviate eye strain. (E)"}\newline
  \textit{"Keeping the eyes healthy is one thing I really care about, so I like this advice. (W12)"}\\
\addlinespace
Contextual information &
  \textit{"The afternoon is a good time for studying. Don’t spend too much time on your phone. (E)"}\newline
  \textit{"Afternoon is indeed my study time during which I should improve my efficiency. It is right. (W4)"}\\
\hline
\multicolumn{2}{c}{\textbf{Semi-structured interview study}} \\
\hline
\textbf{Mental states related to habitual use} & \textbf{Representative quote(s)} \\
\hline
Boredom &
  \textit{"I find myself instinctively reaching for my phone in search of mental stimulation when doing simple assignments light on cognitive engagement." (S1)}\\
\addlinespace
Stress &
  \textit{"One day, work wasn't progressing well and I was so frustrated that I unlocked my phone for a quick view to ease my mood." (S9)}\\
\addlinespace
Inertia &
  \textit{"Upon returning home after work, I intended to transition back into a focused state for reading or other activities but struggled to shift from a relaxed state. At that point, my phone was the tool for procrastination." (S2)}\\
\hline
\textbf{Activity engagement states} & \textbf{Representative quote(s)} \\
\hline
Engaging in activities &
  \textit{"I was reluctant to start handling this challenging work that I scrolled my phone screen anxiously." (S2)}\\
\addlinespace
Not engaging in activities &
  \textit{"After getting off work and returning home, I collapse on the sofa and binge-watch Tiktok for one to two hours." (S9)}\\
\bottomrule
\end{tabular}
\end{table*}

\newpage
\begin{figure*}[b]
    \centering
    \includegraphics[width=\linewidth]{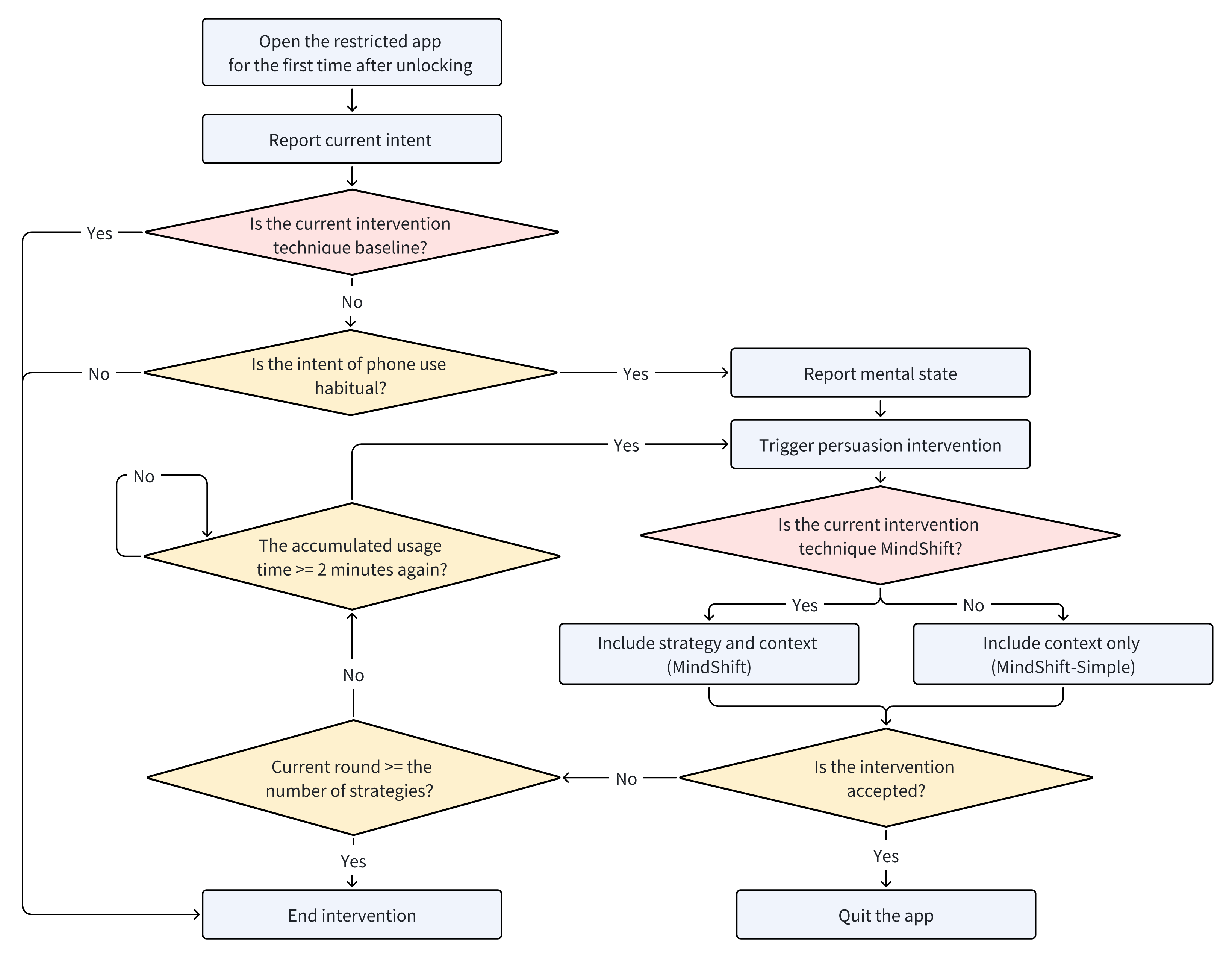}
    \caption{\review{Interaction Process of Three Intervention Techniques. The two red process blocks illustrate the differences between the three intervention techniques.}}
    \label{fig:comparision-tech}
    \Description{This flowchart outlines the interaction process of three intervention techniques. Red process blocks illustrate the differences between the three intervention techniques. It starts with the user opening a restricted app and reporting their intent. Then comes to a red block to determine whether it is baseline. If the intervention technique is not the baseline, the user's habitual intent is assessed. If the intent is habitual, a persuasion intervention is triggered. Then comes to a red block to determine whether it is MindShift. Depending on whether the technique is MindShift or a simpler version, the strategy and context or just the context are included. The intervention continues based on acceptance: if accepted, the user is prompted to quit the app; if not, the process checks if the accumulated usage time is over 2 minutes or if the current round has reached the number of strategies planned. If neither condition is met, the intervention ends.}
\end{figure*}


\end{document}